\documentclass{article}

\usepackage[final]{neurips_2025}

\usepackage[utf8]{inputenc} 
\usepackage[T1]{fontenc}    
\usepackage{hyperref}       
\usepackage{url}            
\usepackage{booktabs}       
\usepackage{amsfonts}       
\usepackage{nicefrac}       
\usepackage{microtype}      
\usepackage{xcolor}         
\usepackage{booktabs}
\usepackage{array}

\usepackage{tcolorbox}
\tcbuselibrary{listingsutf8}
\usepackage{listings}
\usepackage{xcolor}
\usepackage{tabularx}
\usepackage{multirow}
\usepackage{amsmath}
\usepackage{subcaption}

\usepackage{siunitx}
\title{DetectiumFire: A Comprehensive Multi-modal Dataset Bridging Vision and Language for Fire Understanding}

\author{%
 Zixuan Liu \\
  Department of Computer Science \\
  Tulane University \\
   \texttt{zliu41@tulane.edu} \\ 
 \And 
   Siavash Khajavi  \\
   Department of Industrial Engineering and Management \\
   Aalto University \\
   \texttt{siavash.khajavi@aalto.fi} \\
   \And
   Guangkai Jiang \\
   \texttt{guangkaijiang@gmail.com}  \\
}

\begin{document}

\maketitle

\begin{abstract}
  Recent advances in multi-modal models have demonstrated strong performance in tasks such as image generation and reasoning. However, applying these models to the fire domain remains challenging due to the lack of publicly available datasets with high-quality fire domain annotations.  To address this gap, we introduce \textbf{DetectiumFire}, a large-scale, multi-modal dataset comprising of 22.5k high-resolution fire-related images and 2.5k real-world fire-related videos covering a wide range of fire types, environments, and risk levels. The data are annotated with both traditional computer vision labels (e.g., bounding boxes) and detailed textual prompts describing the scene, enabling applications such as synthetic data generation and fire risk reasoning. DetectiumFire offers clear advantages over existing benchmarks in scale, diversity, and data quality, significantly reducing redundancy and enhancing coverage of real-world scenarios. We validate the utility of DetectiumFire across multiple tasks, including object detection, diffusion-based image generation, and vision-language reasoning. Our results highlight the potential of this dataset to advance fire-related research and support the development of intelligent safety systems. We release DetectiumFire to promote broader exploration of fire understanding in the AI community. The dataset is available at \url{ https://kaggle.com/datasets/38b79c344bdfc55d1eed3d22fbaa9c31fad45e27edbbe9e3c529d6e5c4f93890}. 
\end{abstract}

\section{Introduction}
Recent advances in multi-modal models, such as CLIP~\cite{radford2021learning}, Stable Diffusion~\cite{rombach2022high,podell2023sdxl}, and Large Vison Language Model~\cite{zhu2023minigpt,bai2023qwen, li2022blip,wang2024qwen2} have demonstrated remarkable success across a wide range of tasks, including image generation~\cite{rombach2022high,ramesh2021zero}, visual question answering~\cite{xu2024llavacot,li2022blip,chen2023minigpt,zhang2021vinvl}, cross-modal retrieval~\cite{radford2021learning,li2022blip}, and multi-modal reasoning~\cite{xu2024llavacot,zhang2023multimodal,li2025imagine}. Despite their transformational impact, little attention has been given to the field of fire understanding. Existing datasets in this domain are limited to traditional computer vision tasks, such as image classification and object detection~\cite{de2022automatic,phylake1337_fire_dataset,cair_fire_detection,olafenwa_firenet,dincer_wildfire_dataset}, which are outdated and inadequate for training or evaluating modern multi-modal models. This lack of attention can largely be attributed to the challenges of creating large, open, and high-quality datasets for fire understanding, as such endeavors are often restricted to private companies and remain inaccessible to the broader research community. 

To fill this gap, we present \textbf{DetectiumFire}, a publicly available dataset comprising over 22.5k fire-related images and 2.5k fire-related videos, annotated for both conventional computer vision tasks (e.g., classification, object detection) and modern multi-modal applications (e.g., image generation, vision-language reasoning). Our motivation for building this multi-modal dataset goes beyond conventional fire detection tasks and is driven by two core objectives: 

\begin{enumerate}
\item \textbf{Addressing the scarcity and diversity gap in fire datasets.} Although fire safety is a globally critical issue with direct implications for public safety, infrastructure, and climate resilience~\cite{nfpa2025resilience,arcadis2024fire,firerescue2023climate}, collecting high-quality fire data remains extremely challenging. Like in other safety-critical domains such as medical imaging~\cite{sizikova2024synthetic,erickson2017machine} or autonomous vehicles~\cite{song2023synthetic,wong2020mapping}, fire events are rare and often dangerous to capture, resulting in limited publicly available datasets. For instance, the largest existing fire dataset, D-Fire~\cite{de2022automatic}, includes only 5k fire images annotated with bounding boxes, and no large-scale video dataset currently exists for fire-related scenes. In this context, generative models, such as diffusion models, are increasingly adopted for augmenting rare data scenarios (e.g., rare disease imaging)~\cite{hasani2022artificial,zhang2024data,mendes2025synthetic}. Therefore, in addition to solely collecting more real-world fire data, we also aim at building a multi-modal dataset capable of utilizing modern diffusion models to generate high-quality synthetic fire data to address the data scarcity and diversity problem.

    \item \textbf{Enabling modern multi-modal models to reason about fire.} Fire understanding requires more than just detection, it demands contextual reasoning about what is burning, the environment, and the associated risk. Recent advances in multi-modal models, especially vision-language models (VLMs), have shown promise in reducing false positives in fire detection~\cite{gragnaniello2025video,kimintegrated}. However, real-world scenarios often call for finer-grained distinctions. For example, a small candle flame in a safe setting should not raise concern, but the same flame spreading to curtains poses a serious hazard. Existing datasets lack the diversity and annotation detail necessary to support such nuanced reasoning. DetectiumFire addresses this by including a wide range of fire scenarios: both controlled and uncontrolled, each paired with rich captions that describe fire presence, affected objects, environment, and severity. This enables VLMs to learn critical distinctions for effective fire scene interpretation and decision-making in safety-critical applications.
\end{enumerate}

Importantly, our dataset was curated by fire safety professionals with domain expertise and familiarity with AI concepts, ensuring high annotation quality and meaningful scenario coverage. In contrast, existing datasets often lack domain-specific annotations or focus narrowly on detection without considering broader tasks.

Equipped with this comprehensive dataset, we demonstrate how DetectiumFire addresses the data scarcity challenge through diffusion-based generation and how it enables modern vision-language models (VLMs) to reason about fire. We begin by evaluating the advantages of DetectiumFire over the previous largest benchmark, D-Fire~\cite{de2022automatic}, across multiple dimensions. First, we show that DetectiumFire offers broader real-world fire coverage, spanning both indoor and outdoor environments and encompassing a wide range of causes, scales, and risk levels, including many rare or specialized categories absent in prior datasets. Next, we assess dataset quality in terms of redundancy. Using the \texttt{imagededup} tool~\cite{idealods2019imagededup}, we show that DetectiumFire substantially reduces image duplication, with a duplication rate of 0.23 compared to 0.55 in D-Fire. Finally, we evaluate the impact of DetectiumFire on traditional vision tasks such as object detection. Models trained on DetectiumFire significantly generalize better across different benchmarks and achieve improved performance compared to those trained on D-Fire.

To examine the value of synthetic data in addressing the fire data scarcity problem, we investigate two fine-tuning strategies for diffusion models: Supervised Fine-Tuning (SFT) and Reinforcement Learning from Human Feedback (RLHF), and assess the resulting image quality using GPT-4o~\cite{openai2023gpt4}. Our results show that fine-tuning with DetectiumFire leads to substantial improvements in visual fidelity, realism, and prompt alignment. Moreover, these improvements translate into stronger performance on downstream tasks such as object detection. In contrast, models trained on synthetic datasets with lower visual fidelity, such as SD\_Flame~\cite{wang2024flame}, perform significantly worse. These findings validate the usefulness of our synthetic data and highlight DetectiumFire’s potential to serve as a new standard for fire image generation, offering practical advantages over existing methods such as 3D simulation~\cite{borkman2021unity,khajavi2024synthetic} and alternative generative approaches~\cite{xie2020generating}.

Lastly, we assess DetectiumFire’s potential for enabling fire reasoning in modern multi-modal models. To this end, we fine-tune the LLaMA-3.2-11B-Vision-Instruct model~\cite{meta2024llama} to infer three key fire-related attributes from images: the burning object, the surrounding environment, and the severity of the fire. The fine-tuned model achieves strong accuracy across all three tasks, demonstrating its ability to extract meaningful, domain-specific information from DetectiumFire. This capability supports a range of real-world applications, including automated alarm filtering~\cite{scylla2024false}, post-incident analysis~\cite{article}, and early-stage risk assessment~\cite{wtw2023robots}.

In summary, DetectiumFire fills a longstanding gap by offering a large-scale, richly annotated, and multi-modal dataset for fire understanding. It enables advances in both discriminative and generative modeling, and serves as a foundation for reasoning-centered applications critical to public safety. We hope this resource inspires new research directions at the intersection of vision, language, and safety-critical decision-making. The methodology and insights developed here are also transferable to adjacent domains such as disaster response~\cite{camps2024ai}, industrial monitoring, and rare event modeling~\cite{shyalika2024comprehensive}, where data scarcity, domain complexity, and high stakes similarly intersect. We believe this work contributes meaningfully to both applied AI and real-world safety-critical research.

\section{Related Work}
\textbf{Fire Detection Datasets. }
The collection of real-world fire images for computer vision tasks such as classification and object detection has been an active area of research.~\cite{de2022automatic,phylake1337_fire_dataset,cair_fire_detection,olafenwa_firenet,dincer_wildfire_dataset}. The largest publicly available dataset for fire detection to date is the D-Fire~\cite{de2022automatic}, which includes 5,822 images of fire annotated with YOLO-format bounding boxes and 9,838 non-fire images. However, a significant portion of this dataset consists of frames extracted from videos, leading to highly similar and repetitive images. After removing duplicate and near-duplicate images, only half of the dataset remains, limiting the diversity of fire scenarios represented. In contrast, \textbf{DetectiumFire} offers a more comprehensive dataset with over 22.5k fire-related images and 2.5k real-world fire-related videos, encompassing a diverse range of fire-related scenarios and eliminating duplications.

\textbf{Synthetic Fire Image Generation. } Advancements in generative models have facilitated synthetic fire image generation. The FLAME Diffuser~\cite{wang2024flame} is a diffusion-based framework designed to synthesize realistic wildfire images with precise flame location control. This model employs augmented masks, sampled from real wildfire data, and applies Perlin noise to guide the generation of realistic flames. However, FLAME Diffuser is a training-free framework and does not leverage large-scale real-world fire datasets, which may limit the realism and contextual accuracy of the generated images. In contrast, our synthetic data is generated using diffusion models fine-tuned with two distinct pipelines: Supervised Fine-Tuning (SFT) and Reinforcement Learning from Human Feedback (RLHF) on our large-scale text-to-image real-world fire dataset. This approach enables the generation of fire images that are not only more realistic but also semantically coherent, accurately reflecting real-world fire scenarios.

\textbf{Fire Visual Reasoning Benchmarks. }
In the realm of fire visual reasoning, the HAZARD benchmark~\cite{zhou2024hazard} has been proposed to assess the decision-making abilities of AI agents in dynamic situations involving hazards like fire, flood, and wind. HAZARD operates entirely within a simulation environment (TDW), where fire is rendered procedurally with limited realism. While valuable for evaluating decision-making in controlled settings, the lack of real-world data limits its applicability to real-world scenarios. In contrast, DetectiumFire provides a large-scale, diverse real-world dataset with precise annotations, including environmental context, burning objects, and risk levels. This real data is significantly more useful for training deployable models in real-world safety-critical applications compared to virtual images. Moreover, our focus on fire is a deliberate and necessary step toward addressing one of the most high-impact, under-resourced problems in the AI for disaster safety community.

To the best of our knowledge, \textbf{DetectiumFire} is the largest publicly available real-world fire dataset, containing both images and videos. It is the first dataset to include image-to-text annotations with detailed fire domain knowledge specifically designed for multi-modal model training and evaluation.

\section{DetectiumFire Dataset}
\label{sec:data}

DetectiumFire is a comprehensive dataset comprising over 14.5k high-quality real-world fire images and 2.5k fire-related videos. In addition to real-world data, it includes 8k synthetic fire images generated using diffusion-based models, along with 12k preference pairs curated during the RLHF process to enhance model alignment, both of which, as detailed in Section~\ref{sec:syn}, contribute to addressing data scarcity in fire-related applications. Some examples of the dataset can be found in Appendix~\ref{sec:add_example}.

\subsection{Real-World Dataset}
\label{sec:real}
Each sample in this branch of the dataset is annotated with bounding boxes and accompanied by detailed prompts describing fire-related content. To ensure the dataset's quality and utility, we implemented a rigorous data processing pipeline consisting of: (1) data sourcing, (2) quality analysis and filtering, (3) annotation, and (4) human verification. All annotations and verifications were performed by experts in fire-related fields proficient in computer vision and AI techniques.

\textbf{Data Collection. } To assemble a diverse collection of fire data, we conducted extensive web searches for relevant images, videos, and short clips across multiple platforms, including Google, Twitter, YouTube, and TikTok. Text-based queries were employed to retrieve data related to fire incidents and common fire-related scenarios, such as cooking, campfires, burning candles, and incense. While English was primarily used for searches, these platforms index fire-related visual content from around the world, encompassing a wide range of regions and event types, thereby mitigating potential geographic biases. To further enhance comprehensiveness and cultural coverage, we also employed multilingual search strategies, incorporating languages such as Chinese, which is among the most widely spoken languages globally~\cite{wikipedia_languages} and is associated with large and active online communities across the aforementioned platforms. Future work should explore additional languages to further diversify the dataset. We also include some images collected using IoT devices during controlled fire demonstrations. Moreover, we integrated fire samples from several existing benchmarks. A detailed breakdown of all data sources, including licensing information and any modifications performed on each dataset, is provided in Appendix~\ref{sec:source}. We fully acknowledge the contributions of prior datasets and ensure adherence to all stated licenses. 

\textbf{Preprocessing. } To construct the image dataset, we aggregated all collected images and extracted individual frames from the videos and short clips containing fire scenes. To ensure quality and uniqueness, we removed exact duplicates and near-duplicates using the \texttt{imagededup} tool~\cite{idealods2019imagededup}. Subsequently, our human annotators manually evaluated the remaining images, ensuring they provided meaningful fire-related information. Images deemed low-quality or irrelevant were removed during this stage. For the video dataset, human annotators meticulously reviewed each video and short clip to confirm the presence of fire scenes. They assessed whether the fire was clearly identifiable and excluded videos or clips that were of low quality or lacked significant fire-related content.

\textbf{Annotation and Verification. } Following preprocessing, our human annotators generated bounding box annotations for the filtered fire images using the Roboflow platform~\cite{roboflow2024}. This annotation process, which spanned several months, was carefully conducted to ensure the resulting annotations were accurate, comprehensive, and consistent across the dataset. Further information on annotation quality and inter-annotator agreement is available in Appendix~\ref{sec:anno_quality}. After completing this step, we curated a final set of 7k high-quality fire images with 7k non-fire images, providing a contrast for fire recognition tasks. For the video dataset, annotators extracted clips containing identifiable fire scenes. Each extracted clip was required to be at least 10 seconds long to ensure sufficient context. This process resulted in 1.7k video clips featuring fire and an additional 757 non-fire video clips, which were included to facilitate more robust model training and evaluation.

To construct fire-related prompts for each sample, we first use GPT-4o~\cite{openai2023gpt4} to generate detailed descriptions that capture the burning objects, surrounding environment, and fire severity. The prompt templates used to query GPT-4o, along with the rationale for leveraging GPT models in the annotation process, are detailed in Appendix~\ref{sec:caption}. All generated captions are then reviewed and manually edited by human annotators. To streamline this process, we developed a custom Python annotation tool, described in Appendix~\ref{sec:anno_tool}, to assist with verification and refinement. During this step, annotators validated the correctness of each caption, removed irrelevant or misleading content, and revised phrasing to ensure alignment with our fire-focused data generation objectives. In particular, fire severity labels were manually corrected based on a four-level taxonomy: No Risk (e.g. candle flame or contained stove fire), Low Risk (requires continued attention, but not immediate action), Moderate Risk (spreading fire needing action), and High Risk (uncontrolled fires requiring immediate intervention). Because these captions serve as input to diffusion models for synthetic data generation, one of our key goals was to ensure focus and consistency. We observed that longer prompts containing unrelated scene-level details, such as human actions or complex backgrounds, often caused diffusion models to diverge from rendering accurate fire phenomena. To mitigate this, we limit captions to 75 tokens, emphasizing the key aspects: what is burning, the surrounding environment, and the fire’s severity or stage. Despite their brevity, these captions effectively convey the core attributes needed for high-quality fire image synthesis and multi-modal reasoning. In future work, we plan to incorporate richer scene-level annotations that support more advanced reasoning tasks, such as modeling fire progression, assessing human presence and safety risk, and generating multi-step incident reports for intelligent fire response systems.

\textbf{Dataset Diversity and Real-World Fire Coverage. } The final real-world portion of \textbf{DetectiumFire} dataset covers a wide range of fire domains and scenarios, making it significantly more general than previous datasets.  It includes both indoor and outdoor scenes, covering a wide spectrum of fire causes, scales, and severity levels, including many \textbf{rare or underrepresented categories} that are often overlooked in prior work. This diversity is critical for enabling models to generalize across real-world fire detection tasks.  Specifically, the dataset is divided into two major categories:
\textbf{Indoor Fires (3,374 images)} include: Cooking fire (controlled, 54), Stove/fireplace fire (30), Candle flame (1,076), Lighter flame (1,075), Matches flame (163), Kitchen fire (uncontrolled, 212), Electrical fires (e.g., phones, fans, wire; 213), and Other indoor accidental fires (551).
\textbf{Outdoor Fires (4,175 images)} include: Campfires (controlled, 347), Vehicle fires (750), Forest/wildfires (1,045), House/residential fires (1,159), Ship fires (42), Plane fires (14), Trash bin fires (26), LPG/gas tank fires (159), Burning debris such as leaves or woodpiles (190), Other outdoor fires (418), and Pure flame crops (25). A full breakdown is provided in Appendix~\ref{sec:tax}.
This taxonomy highlights that DetectiumFire is not a reassembly of prior resources but the result of extensive efforts, scene filtering, and targeted domain coverage. 
Crucially, unlike most existing datasets that focus exclusively on destructive fire events, DetectiumFire includes a range of \textit{controlled fire scenarios}, such as campfires, stove-top flames, and candlelight, that are common in daily life and should not trigger alarms. This distinction is essential for developing fire detection systems that can minimize false positives. By including both controlled (low-risk) and uncontrolled (high-risk) fires, our dataset supports training models that can reason about fire severity and improve real-world reliability.

As for non-fire images, we go beyond simply adding visually irrelevant content. Instead, we deliberately curate challenging negative examples that are known to frequently trigger false positives in existing fire detection models. These include images of fire-like clouds, sunsets with intense red hues, and scenes with strong light sources in dark environments, all of which may resemble fire under certain conditions. Representative examples are provided in Appendix~\ref{sec:add_example}. These design choices enhance the dataset’s realism and difficulty, ensuring that DetectiumFire not only supports model training but also provides a robust benchmark for evaluating false alarm resistance. Together, these decisions make DetectiumFire more diverse, realistic, and practically useful than previous datasets in this domain.

\subsection{Synthetic Dataset}
\label{sec:syn}

To investigate whether modern generative models can effectively address data scarcity in the fire domain, we explore two widely adopted fine-tuning approaches: (1) Supervised Fine-Tuning (SFT) and (2) Reinforcement Learning from Human Feedback (RLHF). Although these methods are standard, their application to safety-critical and data-scarce domains like fire understanding yields valuable insights into model generalization and the role of high-quality synthetic data in real-world AI systems.

\textbf{Supervised Fine-Tuning. } We fine-tuned multiple widely used pre-trained diffusion models, including Stable Diffusion v1.5, Stable Diffusion 2~\cite{Rombach_2022_CVPR} and Stabe Diffusion XL-1.0~\cite{podell2023sdxl}, on the text-image dataset derived in Section~\ref{sec:real}. For efficient and lightweight training, we adopted LoRA~\cite{hu2021lora} techniques\footnote{Implementation from \url{https://github.com/kohya-ss/sd-scripts.git}}. All models were trained for 4,000 steps with a learning rate of $1e-4$, a prior loss weight of 1, an AdamW8bit optimizer, and mixed precision set to fp16, while keeping all other parameters at their default values. Hyperparameter optimization was not performed, leaving room for potential improvements in performance.

\textbf{Reinforcement Learning from Human Feedback. } Reinforcement Learning from Human Feedback (RLHF) has proven effective in improving multi-modal text-to-image models~\cite{lee2023aligning,fan2024reinforcement,wallace2024diffusion}. To align the generated images with fire-related preferences, we incorporated an RLHF pipeline into this work. 

To construct a preference dataset for the RLHF pipeline, we began by randomly selecting 4k unique prompts from our text-to-image dataset. For each prompt, images were generated using the three diffusion models fine-tuned in the previous step. These generated images were then sent to annotators for labeling. Annotators meticulously reviewed the images and filtered out invalid preference pairs to ensure high-quality annotations. Each prompt receives between $k=2 \sim 9$ unique images, leading to $C_2^k=k(k-1)/2$ preference pairs. For simplicity, we collected binary feedback from multiple annotators in this work, though more detailed feedback, such as ranking, may further improve results~\cite{stiennon2020learning}. Each annotator was presented with two images generated from the same prompt and was tasked with selecting a preference based on the following criteria similar to~\cite{wallace2024diffusion}:
\begin{enumerate}
\item \textbf{General Preference:} How visually appealing and convincing the generated fire appears overall. \item \textbf{Visual Appeal:} The artistic quality, realism, and aesthetic beauty of the generated fire. \item \textbf{Prompt Alignment:} How accurately the generated fire matches the intent and details described in the provided prompt.
\end{enumerate}

Notice that we specifically focused on evaluating the quality of fire generation while ignoring other elements in the image, such as humans or unrelated objects. 
After removing ties, this process resulted in a preference dataset comprising 12k labeled pairs: $D=\{c, x_i^w, x_i^l\}_{i=1}^{N}$, where each example contains a prompt c and a pair of images with human preference $x_i^w  \succ x_i^l$. 

Next, we applied the Diffusion-DPO pipeline described in~\cite{wallace2024diffusion} to fine-tune the Stable Diffusion v1.5 model\footnote{Implementation from \url{https://github.com/SalesforceAIResearch/DiffusionDPO}}. This approach was chosen for two primary reasons. First, Direct Preference Optimization (DPO)~\cite{rafailov2024direct} is a more effective and lightweight supervised method for learning from human preference compared to traditional reinforcement learning techniques. Second, it explicitly avoids the need for reward modeling and instead directly learns preference from the datasets. The model was fine-tuned using the following hyperparameters: maximum training steps set to 3,000, training batch size of 1, gradient accumulation steps of 1, learning rate of $1e-8$ with a constant learning rate scheduler, and a warm-up period of 500 steps. The DPO beta parameter was set to 5,000. Other parameters are set to default. Similar to SFT, we did not perform extensive hyperparameter tuning.

\textbf{Synthetic Data Generation. } To generate fire-related images, we reused prompts from the real-world training dataset. Both the SFT and RLHF fine-tuned models were used to produce synthetic images. We did not perform any hyperparameter tuning and kept the default settings in~\cite{stable-diffusion-webui}.  Human annotators then filtered the generated images based on the criteria outlined in Section~\ref{sec:syn} and added bounding box annotations, following the procedure described in Section~\ref{sec:real}. This process resulted in approximately 8k high-quality synthetic fire images

\section{Experiments}

In this section, we present a comprehensive set of experiments to evaluate the effectiveness of \textbf{DetectiumFire} across a range of tasks. Our goal is to systematically assess the dataset’s value in both traditional computer vision and emerging multi-modal settings. We focus on the following key research questions and leave the additional experiments regarding fire video in Appendix~\ref{sec:video_ex}:

\begin{itemize} \item \textbf{Benchmark Comparison:} How does DetectiumFire compare to existing datasets in terms of diversity, quality, and utility? How well do standard computer vision models perform on DetectiumFire versus prior benchmarks? \item \textbf{Synthetic Data Utility:} Does our synthetic data pipeline produce high-quality fire images? Can this synthetic data improve downstream tasks such as object detection? \item \textbf{Multi-modal Fire Reasoning:} Can DetectiumFire improve the ability of vision-language models (VLMs) to reason about fire-related scenes? \end{itemize}

\subsection{Experimental Details}
\label{sec:ex_details}
\textbf{Benchmark Comparison. } To highlight the advantages of \textbf{DetectiumFire} over the previous largest benchmark dataset, D-Fire~\cite{de2022automatic}, we begin by analyzing the duplication issue, which is common in datasets constructed by extracting frames from videos. This often leads to a high proportion of visually similar or redundant images. To assess and compare duplication levels, we used the \texttt{imagededup} tool~\cite{idealods2019imagededup}, applying two distinct methods: (1) perceptual hashing (PHash) with a maximum Hamming distance threshold of 1, and (2) a CNN-based method using a pretrained EfficientNet-B4 model\footnote{\url{https://pytorch.org/vision/main/models/generated/torchvision.models.efficientnet_b4.html?highlight=efficientnet_b4_weights\#torchvision.models.EfficientNet_B4_Weights}}, with a minimum similarity threshold of 0.9. This dual-method approach allows us to robustly identify both exact and near-duplicate images across the datasets.

To evaluate the performance and generalization ability of object detection models on both datasets, we conducted a 5-fold cross-validation. In each fold, 10\% of the dataset was randomly sampled as a fixed test set, while the remaining 90\% was split into 70\% for training and 20\% for validation. We employed YOLOv11~\cite{yolov11-ultralytics}, the most advanced model in the YOLO family. Specifically, we used the pre-trained YOLOv11m architecture, trained with a mini-batch size of 16 for 300 epochs. All other hyperparameters followed the default settings.

In addition to YOLOv11m, we also evaluated two baseline models on DetectiumFire: Faster R-CNN~\cite{ren2015faster} and YOLO-World-M~\cite{Cheng2024YOLOWorld}. The Faster R-CNN implementation follows the MMDetection framework~\cite{chen2019mmdetection} with a ResNet-101 backbone and Feature Pyramid Network (R-101-FPN), initialized using OpenXLab’s pre-trained weights\footnote{\url{https://download.openxlab.org.cn/models/mmdetection/FasterR-CNN/weight/faster-rcnn_r101-caffe_fpn_1x_coco}}. We set the learning rate to 0.0025 and used a batch size of 2 for 300 epochs. For YOLO-World-M, we followed the official instructions \footnote{\url{https://github.com/AILab-CVC/YOLO-World}}. The category descriptions used for training were extracted from the detailed fire-related captions included in DetectiumFire. All hyperparameters were left at their default settings.
All models are trained on the training set, with the best-performing checkpoint selected based on validation loss. Final results are reported on the held-out test set. All training is conducted using a single NVIDIA Tesla T4 GPU. We report three key metrics to assess detection performance: (1) Mean Average Precision (mAP). (2) Mean Average Precision at 50\% Intersection over Union (mAP\@50IoU), and (3) Mean Average Recall with a maximum of 100 detections (mAR@100).

\textbf{Synthetic Data Utility. } To evaluate the quality and diversity of synthetic data generated by different methods, we use GPT-4o~\cite{openai2023gpt4} to create 770 unique prompts describing diverse fire scenes, which are distinct from those in our dataset captions. These prompts are used to generate images with three models: the original Stable Diffusion v1.5 and two fine-tuned variants introduced in Section~\ref{sec:syn}. While human evaluation is considered the gold standard for generative quality assessment, its high cost and limited scalability motivate the use of automated methods~\cite{dai2023safe}. Following prior work~\cite{chiang2023can}, we adopt GPT-4o as a proxy evaluator. Specifically, GPT-4o compares image pairs generated from the same prompt by different models and scores them along three dimensions introduced in Section~\ref{sec:syn}. The full evaluation prompt is included in Appendix~\ref{sec:gpt_prompt}. To enable a robust and quantitative comparison, we compute Elo scores from pairwise win rates across models, a metric commonly used in LLM evaluations~\cite{askell2021general,dai2023safe,khan2024debating}. Further details about our Elo computation procedure are included in Appendix~\ref{sec:elo}.

To assess whether synthetic data improves downstream performance, we train YOLOv11m and Faster R-CNN models using synthetic datasets generated via both the SFT and RLHF pipelines. As a baseline, we include FLAME\_SD~\cite{wang2024flame}, which contains 10k synthetic fire images generated using a fine-tuned latent diffusion model. We also explore training with combined synthetic (images generated from both SFT and RLHF) and real data to assess the additive benefits of synthetic augmentation.

\textbf{Multi-Modal Fire Reasoning. } To investigate the potential of modern vision-language models (VLMs) in understanding fire-related scenes, we conducted a study using the latest model, LLaMA-3.2-11B-Vision-Instruct~\cite{meta2024llama}. Our goal was to assess whether fine-tuning on DetectiumFire improves the model’s fire reasoning capabilities compared to the original, unmodified model. We fine-tuned the model using our image-caption pairs with LoRA~\cite{hu2022lora} and Fully Sharded Data Parallelism (FSDP)~\cite{zhao2023pytorch}, training on 2× A100 GPUs with a learning rate of 1e-5 for 3 epochs. The fine-tuning process was implemented using the official llama-cookbook repository provided by Meta AI\footnote{\url{https://github.com/meta-llama/llama-cookbook/tree/main}}. We then evaluated the model's ability to infer three key fire-specific attributes from unseen images in the test set: 
\begin{enumerate}
    \item \textbf{Burning Object}: The primary object(s) on fire (e.g., car, tree, stove, building).

    \item \textbf{Surrounding Environment}: The contextual setting of the fire (e.g., indoor, outdoor, industrial, urban, wildland).

    \item\textbf{Fire Severity}: The level of risk posed by the fire, ranging from no immediate concern (e.g., candle flame) to urgent, uncontrolled scenarios (e.g., residential or forest fires).
\end{enumerate}
To measure accuracy, we used GPT-4o~\cite{openai2023gpt4} to evaluate whether the model's outputs matched the ground-truth attributes derived from our human-verified captions.

\subsection{Experiment Results}
\textbf{Benchmark Comparison. } The result for image duplication can be found in Table~\ref{tab:duplication}, which demonstrates that DetectiumFire significantly reduces image duplication compared to D-Fire. This is primarily due to differences in how video frames are sampled and curated. In the case of D-Fire, long fire videos are often processed by extracting every frame without any filtering, resulting in many near-identical images from the same scene. This leads to a large proportion of duplicated or visually redundant images. In contrast, DetectiumFire also uses fire videos, but we apply a careful filtering strategy. For each video, we consider the full fire development process from the early ignition stage to peak intensity, and sample only 1 or 2 visually distinct frames from each key phase. This ensures that even when multiple images come from the same scene, they represent different fire shapes, intensities, or spatial spreads. As a result, our dataset avoids excessive redundancy while still capturing temporal progression. Even among similar samples, the fire appearance is visually diverse, which improves overall dataset quality and utility for training.
\begin{table}[ht]
\centering
\renewcommand{\arraystretch}{1.2}
\setlength{\tabcolsep}{8pt}
\caption{Comparison of image duplication in DetectiumFire and D-Fire using PHash and CNN-based similarity.}
\label{tab:duplication}
\begin{tabular}{@{}l l r r r@{}}
\toprule
\textbf{Method} & \textbf{Dataset} & \textbf{Total Images} & \textbf{Duplicated} & \textbf{Duplication Ratio} \\
\midrule
PHash & DetectiumFire & 7,549 & 205  & 0.03 \\
      & D-Fire        & 5,822 & 883  & 0.15 \\
CNN   & DetectiumFire & 7,549 & 1,756 & 0.23 \\
      & D-Fire        & 5,822 & 3,201 & 0.55 \\
\bottomrule
\end{tabular}
\end{table}

\begin{table}[ht]
\centering
\caption{Cross-dataset object detection performance using YOLOv11m.}
\label{tab:cross_dataset_detection}
\begin{tabular}{@{}llccc@{}}
\toprule
\textbf{Training Set} & \textbf{Test Set} & \textbf{mAP (\%)} & \textbf{mAP@50 (\%)} & \textbf{mAR (\%)} \\
\midrule
DetectiumFire & DetectiumFire & 43.74 ± 0.64 & 75.96 ± 0.62 & 69.80 ± 1.25 \\
D-Fire & D-Fire & 40.28 ± 1.76 & 71.66 ± 2.06 & 65.12 ± 2.66 \\
DetectiumFire & D-Fire & 40.32 ± 1.53 & 68.94 ± 1.43 & 63.72 ± 1.46 \\
D-Fire & DetectiumFire & 24.88 ± 0.77 & 48.04 ± 1.21 & 47.42 ± 1.62 \\
\bottomrule
\end{tabular}%
\end{table}

The performance of YOLOv11m on both datasets is summarized in Table~\ref{tab:cross_dataset_detection}. The results show that models trained on DetectiumFire generalize well to the D-Fire test set, achieving performance comparable to models trained directly on D-Fire. Notably, there is no overlap between the two datasets, which highlights the strong generalization capacity and broad coverage of DetectiumFire. In contrast, models trained on D-Fire perform significantly worse when evaluated on DetectiumFire. This performance gap suggests that D-Fire lacks the diversity and complexity present in DetectiumFire. These findings confirm that DetectiumFire serves as a more comprehensive and challenging benchmark for fire detection, supporting the development of models that are better suited for real-world safety-critical applications.

\textbf{Synthetic Data Utility. }  The results and discussion for Elo scores and additional synthetic image comparisons are provided in Appendix~\ref{sec:elo_result} and~\ref{sec:compare}, demonstrating that both SFT and RLHF leveraging DetectiumFire dataset significantly improve generation quality compared to the original diffusion model. Furthermore, Table~\ref{tab:object} presents detailed object detection results across various training setups. Notably, we observe that YOLOv11m underperforms Faster R-CNN in certain settings—a finding we further analyze in Appendix~\ref{sec:yolo_fast}. In addition, the performance of YOLO-World-M is particularly promising, highlighting the potential of combining open-vocabulary detectors with richly annotated fire datasets. Given that each image in DetectiumFire is paired with a detailed description of the scene, we believe our dataset provides a natural foundation for advancing open-vocabulary fire understanding, particularly in enabling zero-shot fire detection, detecting unseen object-fire combinations, and leveraging text-guided detection in dynamic or ambiguous environments. 

As for synthetic data, we found that models trained on synthetic data generated via the RLHF pipeline show slightly lower performance than those trained on SFT-generated data. This suggests that the RLHF pipeline might have reduced the diversity of the synthetic dataset. On the other hand, the Flame\_SD dataset~\cite{wang2024flame} yields significantly worse results than all other datasets. We believe the performance gap highlights important differences in dataset generation strategy, data quality, and coverage. We included a more detailed discussion in Appendix~\ref{sec:flame_detectium}. However, this result should not be interpreted as a limitation of synthetic data in general. The experiments demonstrate that diffusion models fine-tuned on DetectiumFire, using a standard pipeline (SFT or RLHF), achieve comparable performance to models trained on real images. This provides strong evidence that high-quality synthetic fire data generated with semantically relevant prompts is indeed useful for downstream tasks, validating the utility of synthetic images.

Another interesting observation is that combining real-world and all synthetic data leads to a slight improvement over using real-world data alone. This is particularly noteworthy given that all prompts and images used to generate the synthetic dataset are derived from the real-world training dataset. This indicates that synthetic data acts as an augmentation to the real-world dataset, introducing additional variations and scenarios that enhance model generalization. The marginal improvement suggests that synthetic data complements real-world data effectively, although it remains insufficient on its own to achieve optimal performance. Future work could focus on refining synthetic data generation methods to further improve diversity and contextual realism, maximizing the benefits of combining real and synthetic data.

\begin{table*}[ht]
\caption{Evaluation performance of various models trained on different datasets. Metrics include mAP, mAP@IoU50, and mAR@100 on the DetectiumFire test set.}
\vskip 0.15in
\centering
\begin{sc}
\resizebox{\textwidth}{!}{%
\begin{tabular}{@{}llccc@{}}
\toprule
\multirow{2}{*}{\textbf{Training Data}} & \multirow{2}{*}{\textbf{Model}} & \multicolumn{3}{c}{\textbf{$\mathbf{DetectiumFire_{\text{test}}}$}} \\
\cmidrule(lr){3-5}
& & $mAP (\%)$ & $mAP@IoU50 (\%)$ & $mAR@100 (\%)$ \\
\midrule
\multirow{3}{*}{Real-world train set} 
& YOLOv11m & 43.74$\pm$0.64 & 75.96$\pm$0.62 & 69.80$\pm$1.25 \\
& Faster-RCNN & 41.13$\pm$1.01 & 80.13$\pm$1.51 & 51.60$\pm$0.54 \\
& YOLO-World-M & 41.28$\pm$0.84 & 76.50$\pm$0.90 & 67.32$\pm$1.43 \\
\midrule
\multirow{2}{*}{Synthetic set (SFT)} 
& YOLOv11m & 33.50$\pm$0.35 & 59.40$\pm$0.75 & 60.54$\pm$2.26\\
& Faster-RCNN & 24.32$\pm$0.40 & 51.18$\pm$0.48 & 37.98$\pm$0.64 \\
\midrule
\multirow{2}{*}{Synthetic set (RLHF)} 
& YOLOv11m & 32.12$\pm$0.29 & 58.08$\pm$0.14 & 59.42$\pm$2.35 \\
& Faster-RCNN & 23.76$\pm$0.19 & 50.31$\pm$0.33 & 36.82$\pm$0.40 \\
\midrule
\multirow{2}{*}{FLAME\_SD} 
& YOLOv11m & 2.10$\pm$3.90 & 6.71$\pm$2.37 & 9.38$\pm$2.83 \\
& Faster-RCNN & 3.42$\pm$0.24 & 7.11$\pm$0.39 & 10.56$\pm$0.15 \\
\midrule
\multirow{2}{*}{Real + Synthetic} 
& YOLOv11m & 44.52$\pm$0.50 & 76.26$\pm$0.34 & 69.92$\pm$1.95 \\
& Faster-RCNN & 42.56$\pm$2.44 & 81.41$\pm$1.93 & 52.18$\pm$0.47 \\
\bottomrule
\end{tabular}%
}
\end{sc}
\label{tab:object}
\vskip -0.1in
\end{table*}

\textbf{Multi-Modal Fire Reasoning. } The results are shown in Table~\ref{tab:vlm_reasoning}, which clearly demonstrate the value of fine-tuning vision-language models (VLMs) on domain-specific data. Additional qualitative analyses are provided in Appendix~\ref{sec:vlm_reasoning}. Across all three reasoning tasks, the model fine-tuned on DetectiumFire outperforms the base LLaMA-3.2-11B-Vision-Instruct by a large margin. The improvement is especially pronounced in severity classification, where accuracy increases from 56.06\% to 83.84\%, indicating that the model becomes substantially more capable of interpreting risk levels from visual clues. Similarly, accuracy in identifying the environment improves by over 17 percentage points, while performance on burning object recognition improves by nearly 25 percentage points. These gains suggest that the curated image-caption pairs in DetectiumFire provide strong supervision signals that enhance the model's capacity to extract and reason about fire-specific visual information, crucial for applications in automated fire assessment, alarm filtering, and intelligent decision support. 

\begin{table}[ht]
\centering
\caption{Accuracy of VLM predictions on fire reasoning tasks before and after fine-tuning on DetectiumFire.}
\label{tab:vlm_reasoning}
\resizebox{\linewidth}{!}{%
\begin{tabular}{lccc}
\toprule
\textbf{Model} & \textbf{Burning Object} & \textbf{Environment} & \textbf{Fire Severity} \\
\midrule
LLaMA-3.2-11B-Vision-Instruct (Base) & 62.73\% & 71.82\% & 56.06\% \\
+ Our Dataset (SFT) & \textbf{87.37\%} & \textbf{89.39\%} & \textbf{83.84\%} \\
\bottomrule
\end{tabular}
}
\end{table}

\section{Conclusion}
\label{sec:conclustion}

In this work, we introduce \textbf{DetectiumFire}, the first large-scale, multi-modal dataset dedicated to comprehensive fire understanding. Our dataset addresses critical gaps in existing benchmarks by offering a diverse, high-quality collection of over 14.5k real-world fire images, 2.5k annotated fire videos, and 8k synthetically generated images with accompanying RLHF preference data. All samples are richly annotated with captions, bounding boxes, and scene-level metadata, enabling a broad range of tasks from conventional object detection to advanced multi-modal reasoning.

Through extensive experiments, we demonstrate the strong generalization ability of models trained on DetectiumFire, outperforming previous benchmarks. We further show that fine-tuning generative models with our dataset significantly improves the visual quality and semantic alignment of synthetic fire imagery, which in turn benefits downstream tasks like object detection. Additionally, we showcase DetectiumFire’s ability to improve the fire reasoning capabilities of vision-language models (VLMs), enabling them to infer critical scene properties such as fire severity, context, and affected objects. These findings highlight the potential of our dataset not only as a benchmark but as a foundation for developing AI systems capable of safe, reliable, and context-aware fire understanding.

We release DetectiumFire in the hope of supporting further research in fire detection, safety-critical AI, synthetic data generation, and vision-language reasoning. We believe our contributions will inspire new directions in areas such as controllable fire video generation, fine-grained fire assessment, and AI agent-based safety response systems, ultimately driving progress toward more resilient and intelligent disaster management technologies.

\section{Acknowledgments}
We sincerely thank Google and Microsoft for their generous cloud credits provided to the Detectium startup. We are grateful to Detectium for sharing valuable data and to EIT Digital for their support of the Detectium startup. We also acknowledge the insightful guidance and advice of Professors Jan Holmström, Kary Främling and Zizhan Zheng. We are indebted to Shuzhen Qi and Jixiang Li for assisting with data collection and offering helpful feedback. Finally, we thank the anonymous reviewers for their constructive comments, which substantially improved the clarity, depth, and presentation of this paper.

{
\small
\bibliographystyle{unsrt}
\bibliography{neurips_2025}

\begin{thebibliography}{10}

\bibitem{radford2021learning}
Alec Radford, Jong~Wook Kim, Chris Hallacy, Aditya Ramesh, Gabriel Goh, Sandhini Agarwal, Girish Sastry, Amanda Askell, Pamela Mishkin, Jack Clark, et~al.
\newblock Learning transferable visual models from natural language supervision.
\newblock In {\em International conference on machine learning}, pages 8748--8763. PMLR, 2021.

\bibitem{rombach2022high}
Robin Rombach, Andreas Blattmann, Dominik Lorenz, Patrick Esser, and Bj{\"o}rn Ommer.
\newblock High-resolution image synthesis with latent diffusion models.
\newblock In {\em Proceedings of the IEEE/CVF conference on computer vision and pattern recognition}, pages 10684--10695, 2022.

\bibitem{podell2023sdxl}
Dustin Podell, Zion English, Kyle Lacey, Andreas Blattmann, Tim Dockhorn, Jonas M{\"u}ller, Joe Penna, and Robin Rombach.
\newblock Sdxl: Improving latent diffusion models for high-resolution image synthesis.
\newblock {\em arXiv preprint arXiv:2307.01952}, 2023.

\bibitem{zhu2023minigpt}
Deyao Zhu, Jun Chen, Xiaoqian Shen, Xiang Li, and Mohamed Elhoseiny.
\newblock Minigpt-4: Enhancing vision-language understanding with advanced large language models.
\newblock {\em arXiv preprint arXiv:2304.10592}, 2023.

\bibitem{bai2023qwen}
Jinze Bai, Shuai Bai, Shusheng Yang, Shijie Wang, Sinan Tan, Peng Wang, Junyang Lin, Chang Zhou, and Jingren Zhou.
\newblock Qwen-vl: A frontier large vision-language model with versatile abilities.
\newblock {\em arXiv preprint arXiv:2308.12966}, 2023.

\bibitem{li2022blip}
Junnan Li, Dongxu Li, Caiming Xiong, and Steven Hoi.
\newblock Blip: Bootstrapping language-image pre-training for unified vision-language understanding and generation.
\newblock In {\em International Conference on Machine Learning (ICML)}, 2022.

\bibitem{wang2024qwen2}
Peng Wang, Shuai Bai, Sinan Tan, Shijie Wang, Zhihao Fan, Jinze Bai, Keqin Chen, Xuejing Liu, Jialin Wang, Wenbin Ge, et~al.
\newblock Qwen2-vl: Enhancing vision-language model's perception of the world at any resolution.
\newblock {\em arXiv preprint arXiv:2409.12191}, 2024.

\bibitem{ramesh2021zero}
Aditya Ramesh, Mikhail Pavlov, Gabriel Goh, Scott Gray, Chelsea Voss, Alec Radford, Mark Chen, and Ilya Sutskever.
\newblock Zero-shot text-to-image generation.
\newblock In {\em International conference on machine learning}, pages 8821--8831. Pmlr, 2021.

\bibitem{xu2024llavacot}
Guowei Xu, Peng Jin, Hao Li, Yibing Song, Lichao Sun, and Li~Yuan.
\newblock Llava-cot: Let vision language models reason step-by-step, 2024.

\bibitem{chen2023minigpt}
Jun Chen, Deyao Zhu, Xiaoqian Shen, Xiang Li, Zechun Liu, Pengchuan Zhang, Raghuraman Krishnamoorthi, Vikas Chandra, Yunyang Xiong, and Mohamed Elhoseiny.
\newblock Minigpt-v2: large language model as a unified interface for vision-language multi-task learning.
\newblock {\em arXiv preprint arXiv:2310.09478}, 2023.

\bibitem{zhang2021vinvl}
Pengchuan Zhang, Xiujun Li, Xiaowei Hu, Jianwei Yang, Lei Zhang, Lijuan Wang, Yejin Choi, and Jianfeng Gao.
\newblock Vinvl: Revisiting visual representations in vision-language models.
\newblock In {\em Proceedings of the IEEE/CVF conference on computer vision and pattern recognition}, pages 5579--5588, 2021.

\bibitem{zhang2023multimodal}
Zhuosheng Zhang, Aston Zhang, Mu~Li, Hai Zhao, George Karypis, and Alex Smola.
\newblock Multimodal chain-of-thought reasoning in language models.
\newblock {\em arXiv preprint arXiv:2302.00923}, 2023.

\bibitem{li2025imagine}
Chengzu Li, Wenshan Wu, Huanyu Zhang, Yan Xia, Shaoguang Mao, Li~Dong, Ivan Vuli{\'c}, and Furu Wei.
\newblock Imagine while reasoning in space: Multimodal visualization-of-thought.
\newblock {\em arXiv preprint arXiv:2501.07542}, 2025.

\bibitem{de2022automatic}
Pedro Vinicius~AB de~Venancio, Adriano~C Lisboa, and Adriano~V Barbosa.
\newblock An automatic fire detection system based on deep convolutional neural networks for low-power, resource-constrained devices.
\newblock {\em Neural Computing and Applications}, 34(18):15349--15368, 2022.

\bibitem{phylake1337_fire_dataset}
Phylake1337.
\newblock Fire dataset.
\newblock \url{https://www.kaggle.com/datasets/phylake1337/fire-dataset}, 2024.
\newblock Accessed: 2024-11-27.

\bibitem{cair_fire_detection}
CAIR Lab.
\newblock Fire-detection-image-dataset.
\newblock \url{https://github.com/cair/Fire-Detection-Image-Dataset}, n.d.
\newblock Accessed: 2024-11-27.

\bibitem{olafenwa_firenet}
Olafenwa Moses.
\newblock Firenet.
\newblock \url{https://github.com/OlafenwaMoses/FireNET}, n.d.
\newblock Accessed: 2024-11-27.

\bibitem{dincer_wildfire_dataset}
Baris Dincer.
\newblock Wildfire detection image data for machine learning process.
\newblock \url{https://www.kaggle.com/brsdincer/wildfire-detection-image-data}, 2021.
\newblock Database Contents License (DbCL) v1.0, Version 1.0.

\bibitem{nfpa2025resilience}
National Fire~Protection Association.
\newblock Resilience revisited: Why fire safety is critical to resilient design.
\newblock {\em NFPA Journal}, 2025.
\newblock Accessed: 2025-05-03.

\bibitem{arcadis2024fire}
Ali Khalid.
\newblock The rising importance of fire safety and protection risks.
\newblock \url{https://www.arcadis.com/en-us/insights/blog/united-states/ali-khalid/2024/the-rising-importance-of-fire-safety-and-protection-risks}, 2024.
\newblock Accessed: 2025-05-03.

\bibitem{firerescue2023climate}
FireRescue1.
\newblock How climate change impacts the fire service.
\newblock \url{https://www.firerescue1.com/risks-and-impact-climate-change-fire-service}, 2023.
\newblock Accessed: 2025-05-03.

\bibitem{sizikova2024synthetic}
Elena Sizikova, Andreu Badal, Jana~G Delfino, Miguel Lago, Brandon Nelson, Niloufar Saharkhiz, Berkman Sahiner, Ghada Zamzmi, and Aldo Badano.
\newblock Synthetic data in radiological imaging: current state and future outlook.
\newblock {\em BJR| Artificial Intelligence}, 1(1):ubae007, 2024.

\bibitem{erickson2017machine}
Bradley~J Erickson, Panagiotis Korfiatis, Zeynettin Akkus, and Timothy~L Kline.
\newblock Machine learning for medical imaging.
\newblock {\em radiographics}, 37(2):505--515, 2017.

\bibitem{song2023synthetic}
Zhihang Song, Zimin He, Xingyu Li, Qiming Ma, Ruibo Ming, Zhiqi Mao, Huaxin Pei, Lihui Peng, Jianming Hu, Danya Yao, et~al.
\newblock Synthetic datasets for autonomous driving: A survey.
\newblock {\em IEEE Transactions on Intelligent Vehicles}, 9(1):1847--1864, 2023.

\bibitem{wong2020mapping}
Kelvin Wong, Yanlei Gu, and Shunsuke Kamijo.
\newblock Mapping for autonomous driving: Opportunities and challenges.
\newblock {\em IEEE Intelligent Transportation Systems Magazine}, 13(1):91--106, 2020.

\bibitem{hasani2022artificial}
Navid Hasani, Faraz Farhadi, Michael~A Morris, Moozhan Nikpanah, Arman Rhamim, Yanji Xu, Anne Pariser, Michael~T Collins, Ronald~M Summers, Elizabeth Jones, et~al.
\newblock Artificial intelligence in medical imaging and its impact on the rare disease community: threats, challenges and opportunities.
\newblock {\em PET clinics}, 17(1):13, 2022.

\bibitem{zhang2024data}
Jiaying Zhang, Guibo Luo, Zi’Ang Zhang, and Yuesheng Zhu.
\newblock Data augmentation in class-conditional diffusion model for semi-supervised medical image segmentation.
\newblock In {\em 2024 International Joint Conference on Neural Networks (IJCNN)}, pages 1--8. IEEE, 2024.

\bibitem{mendes2025synthetic}
Jorge~M Mendes, Aziz Barbar, and Marwa Refaie.
\newblock Synthetic data generation: a privacy-preserving approach to accelerate rare disease research.
\newblock {\em Frontiers in Digital Health}, 7:1563991, 2025.

\bibitem{gragnaniello2025video}
Diego Gragnaniello, Antonio Greco, Carlo Sansone, and Bruno Vento.
\newblock Video fire recognition using zero-shot vision-language models guided by a task-aware object detector.
\newblock {\em ACM Transactions on Multimedia Computing, Communications and Applications}, 2025.

\bibitem{kimintegrated}
Joanne Kim, Yejin Lee, DongSik Yoon, Chansung Jung, and Gunhee Lee.
\newblock An integrated yolo and vlm system for fire detection in enclosed environments.
\newblock In {\em I Can't Believe It's Not Better: Challenges in Applied Deep Learning}.

\bibitem{idealods2019imagededup}
Tanuj Jain, Christopher Lennan, Zubin John, and Dat Tran.
\newblock Imagededup.
\newblock \url{https://github.com/idealo/imagededup}, 2019.

\bibitem{openai2023gpt4}
OpenAI.
\newblock Gpt-4 technical report.
\newblock {\em arXiv preprint arXiv:2303.08774}, 2023.

\bibitem{wang2024flame}
Hao Wang, Sayed Pedram~Haeri Boroujeni, Xiwen Chen, Ashish Bastola, Huayu Li, and Abolfazl Razi.
\newblock Flame diffuser: Grounded wildfire image synthesis using mask guided diffusion.
\newblock {\em arXiv preprint arXiv:2403.03463}, 2024.

\bibitem{borkman2021unity}
Steve Borkman, Adam Crespi, Saurav Dhakad, Sujoy Ganguly, Jonathan Hogins, You-Cyuan Jhang, Mohsen Kamalzadeh, Bowen Li, Steven Leal, Pete Parisi, et~al.
\newblock Unity perception: generate synthetic data for computer vision.
\newblock {\em arXiv preprint arXiv:2107.04259}, 2021.

\bibitem{khajavi2024synthetic}
Siavash~H Khajavi, Mehdi Moshtaghi, Dikai Yu, Zixuan Liu, Kary Fr{\"a}mling, and Jan Holmstr{\"o}m.
\newblock Synthetic imagery for fuzzy object detection: A comparative study.
\newblock {\em arXiv preprint arXiv:2410.01124}, 2024.

\bibitem{xie2020generating}
Chao Xie and Huanjie Tao.
\newblock Generating realistic smoke images with controllable smoke components.
\newblock {\em IEEE Access}, 8:201418--201427, 2020.

\bibitem{meta2024llama}
Meta AI.
\newblock Llama-3.2-11b-vision-instruct.
\newblock \url{https://huggingface.co/meta-llama/Llama-3.2-11B-Vision-Instruct}, 2024.
\newblock Accessed: 2025-05-03.

\bibitem{scylla2024false}
Scylla AI.
\newblock False alarm filtering with ai video analytics.
\newblock \url{https://www.scylla.ai/false-alarm-filtering}, 2024.

\bibitem{article}
Ethan Lee, Isabella Rossi, and Kimberly Jane.
\newblock Post-incident analysis: Ai-driven tools for analyzing fire incidents to improve future safety designs.
\newblock 09 2024.

\bibitem{wtw2023robots}
Willis~Towers Watson.
\newblock Beyond science fiction: The promise of ai-powered risk assessment robots.
\newblock \url{https://www.wtwco.com/en-vn/insights/2023/06/beyond-science-fiction-the-promise-of-ai-powered-risk-assessment-robots}, 2023.

\bibitem{camps2024ai}
Gustau Camps-Valls, Miguel-{\'A}ngel Fern{\'a}ndez-Torres, Kai-Hendrik Cohrs, Adrian H{\"o}hl, Andrea Castelletti, Aytac Pacal, Claire Robin, Francesco Martinuzzi, Ioannis Papoutsis, Ioannis Prapas, et~al.
\newblock Ai for extreme event modeling and understanding: Methodologies and challenges.
\newblock {\em arXiv preprint arXiv:2406.20080}, 2024.

\bibitem{shyalika2024comprehensive}
Chathurangi Shyalika, Ruwan Wickramarachchi, and Amit~P Sheth.
\newblock A comprehensive survey on rare event prediction.
\newblock {\em ACM Computing Surveys}, 57(3):1--39, 2024.

\bibitem{zhou2024hazard}
Qinhong Zhou, Sunli Chen, Yisong Wang, Haozhe Xu, Weihua Du, Hongxin Zhang, Yilun Du, Joshua~B Tenenbaum, and Chuang Gan.
\newblock Hazard challenge: Embodied decision making in dynamically changing environments.
\newblock {\em arXiv preprint arXiv:2401.12975}, 2024.

\bibitem{wikipedia_languages}
{Wikipedia contributors}.
\newblock List of languages by number of native speakers.
\newblock \url{https://en.wikipedia.org/wiki/List_of_languages_by_number_of_native_speakers}, n.d.
\newblock Accessed: 2024-11-27.

\bibitem{roboflow2024}
Brad Dwyer, Joseph Nelson, Trevor Hansen, et~al.
\newblock Roboflow (version 1.0), 2024.

\bibitem{Rombach_2022_CVPR}
Robin Rombach, Andreas Blattmann, Dominik Lorenz, Patrick Esser, and Bj{\"o}rn Ommer.
\newblock High-resolution image synthesis with latent diffusion models.
\newblock In {\em Proceedings of the IEEE/CVF Conference on Computer Vision and Pattern Recognition (CVPR)}, pages 10684--10695, June 2022.

\bibitem{hu2021lora}
Edward~J Hu, Yelong Shen, Phillip Wallis, Zeyuan Allen-Zhu, Yuanzhi Li, Shean Wang, Lu~Wang, and Weizhu Chen.
\newblock Lora: Low-rank adaptation of large language models.
\newblock {\em arXiv preprint arXiv:2106.09685}, 2021.

\bibitem{lee2023aligning}
Kimin Lee, Hao Liu, Moonkyung Ryu, Olivia Watkins, Yuqing Du, Craig Boutilier, Pieter Abbeel, Mohammad Ghavamzadeh, and Shixiang~Shane Gu.
\newblock Aligning text-to-image models using human feedback.
\newblock {\em arXiv preprint arXiv:2302.12192}, 2023.

\bibitem{fan2024reinforcement}
Ying Fan, Olivia Watkins, Yuqing Du, Hao Liu, Moonkyung Ryu, Craig Boutilier, Pieter Abbeel, Mohammad Ghavamzadeh, Kangwook Lee, and Kimin Lee.
\newblock Reinforcement learning for fine-tuning text-to-image diffusion models.
\newblock {\em Advances in Neural Information Processing Systems}, 36, 2024.

\bibitem{wallace2024diffusion}
Bram Wallace, Meihua Dang, Rafael Rafailov, Linqi Zhou, Aaron Lou, Senthil Purushwalkam, Stefano Ermon, Caiming Xiong, Shafiq Joty, and Nikhil Naik.
\newblock Diffusion model alignment using direct preference optimization.
\newblock In {\em Proceedings of the IEEE/CVF Conference on Computer Vision and Pattern Recognition}, pages 8228--8238, 2024.

\bibitem{stiennon2020learning}
Nisan Stiennon, Long Ouyang, Jeffrey Wu, Daniel Ziegler, Ryan Lowe, Chelsea Voss, Alec Radford, Dario Amodei, and Paul~F Christiano.
\newblock Learning to summarize with human feedback.
\newblock {\em Advances in Neural Information Processing Systems}, 33:3008--3021, 2020.

\bibitem{rafailov2024direct}
Rafael Rafailov, Archit Sharma, Eric Mitchell, Christopher~D Manning, Stefano Ermon, and Chelsea Finn.
\newblock Direct preference optimization: Your language model is secretly a reward model.
\newblock {\em Advances in Neural Information Processing Systems}, 36, 2024.

\bibitem{stable-diffusion-webui}
{AUTOMATIC1111}.
\newblock Stable diffusion webui.
\newblock \url{https://github.com/AUTOMATIC1111/stable-diffusion-webui}, 2024.
\newblock Accessed: Dec. 30, 2024.

\bibitem{yolov11-ultralytics}
Ultralytics.
\newblock Yolov11 - ultralytics repository, 2023.
\newblock Accessed: 2024-10-29.

\bibitem{ren2015faster}
Shaoqing Ren, Kaiming He, Ross Girshick, and Jian Sun.
\newblock Faster r-cnn: Towards real-time object detection with region proposal networks.
\newblock {\em Advances in neural information processing systems}, 28, 2015.

\bibitem{Cheng2024YOLOWorld}
Tianheng Cheng, Lin Song, Yixiao Ge, Wenyu Liu, Xinggang Wang, and Ying Shan.
\newblock Yolo-world: Real-time open-vocabulary object detection.
\newblock In {\em Proc. IEEE Conf. Computer Vision and Pattern Recognition (CVPR)}, 2024.

\bibitem{chen2019mmdetection}
Kai Chen, Jiaqi Wang, Jiangmiao Pang, Yuhang Cao, Yu~Xiong, Xiaoxiao Li, Shuyang Sun, Wansen Feng, Ziwei Liu, Jiarui Xu, et~al.
\newblock Mmdetection: Open mmlab detection toolbox and benchmark.
\newblock {\em arXiv preprint arXiv:1906.07155}, 2019.

\bibitem{dai2023safe}
Josef Dai, Xuehai Pan, Ruiyang Sun, Jiaming Ji, Xinbo Xu, Mickel Liu, Yizhou Wang, and Yaodong Yang.
\newblock Safe rlhf: Safe reinforcement learning from human feedback.
\newblock {\em arXiv preprint arXiv:2310.12773}, 2023.

\bibitem{chiang2023can}
Cheng-Han Chiang and Hung-yi Lee.
\newblock Can large language models be an alternative to human evaluations?
\newblock {\em arXiv preprint arXiv:2305.01937}, 2023.

\bibitem{askell2021general}
Amanda Askell, Yuntao Bai, Anna Chen, Dawn Drain, Deep Ganguli, Tom Henighan, Andy Jones, Nicholas Joseph, Ben Mann, Nova DasSarma, et~al.
\newblock A general language assistant as a laboratory for alignment.
\newblock {\em arXiv preprint arXiv:2112.00861}, 2021.

\bibitem{khan2024debating}
Akbir Khan, John Hughes, Dan Valentine, Laura Ruis, Kshitij Sachan, Ansh Radhakrishnan, Edward Grefenstette, Samuel~R Bowman, Tim Rockt{\"a}schel, and Ethan Perez.
\newblock Debating with more persuasive llms leads to more truthful answers.
\newblock {\em arXiv preprint arXiv:2402.06782}, 2024.

\bibitem{hu2022lora}
Edward~J Hu, Yelong Shen, Phillip Wallis, Zeyuan Allen-Zhu, Yuanzhi Li, Shean Wang, Lu~Wang, Weizhu Chen, et~al.
\newblock Lora: Low-rank adaptation of large language models.
\newblock {\em ICLR}, 1(2):3, 2022.

\bibitem{zhao2023pytorch}
Yanli Zhao, Andrew Gu, Rohan Varma, Liang Luo, Chien-Chin Huang, Min Xu, Less Wright, Hamid Shojanazeri, Myle Ott, Sam Shleifer, et~al.
\newblock Pytorch fsdp: experiences on scaling fully sharded data parallel.
\newblock {\em arXiv preprint arXiv:2304.11277}, 2023.

\bibitem{kokosza2024scintilla}
Andrzej Kokosza, Helge Wrede, Daniel Gonzalez~Esparza, Milosz Makowski, Daoming Liu, Dominik~L Michels, Soren Pirk, and Wojtek Palubicki.
\newblock Scintilla: Simulating combustible vegetation for wildfires.
\newblock {\em ACM Transactions on Graphics (TOG)}, 43(4):1--21, 2024.

\bibitem{khajavi2023digital}
Siavash~H Khajavi, M{\"u}ge Tetik, Zixuan Liu, Pasi Korhonen, and Jan Holmstr{\"o}m.
\newblock Digital twin for safety and security: Perspectives on building lifecycle.
\newblock {\em IEEE Access}, 11:52339--52356, 2023.

\bibitem{hou2024assessing}
Wenpin Hou and Zhicheng Ji.
\newblock Assessing gpt-4 for cell type annotation in single-cell rna-seq analysis.
\newblock {\em Nature methods}, 21(8):1462--1465, 2024.

\bibitem{tan2024large}
Zhen Tan, Dawei Li, Song Wang, Alimohammad Beigi, Bohan Jiang, Amrita Bhattacharjee, Mansooreh Karami, Jundong Li, Lu~Cheng, and Huan Liu.
\newblock Large language models for data annotation and synthesis: A survey.
\newblock {\em arXiv preprint arXiv:2402.13446}, 2024.

\bibitem{gberta_2021_ICML}
Gedas Bertasius, Heng Wang, and Lorenzo Torresani.
\newblock Is space-time attention all you need for video understanding?
\newblock In {\em Proceedings of the International Conference on Machine Learning (ICML)}, July 2021.

\bibitem{li2024videomamba}
Kunchang Li, Xinhao Li, Yi~Wang, Yinan He, Yali Wang, Limin Wang, and Yu~Qiao.
\newblock Videomamba: State space model for efficient video understanding.
\newblock In {\em European conference on computer vision}, pages 237--255. Springer, 2024.

\bibitem{kay2017kinetics}
Will Kay, Joao Carreira, Karen Simonyan, Brian Zhang, Chloe Hillier, Sudheendra Vijayanarasimhan, Fabio Viola, Tim Green, Trevor Back, Paul Natsev, et~al.
\newblock The kinetics human action video dataset.
\newblock {\em arXiv preprint arXiv:1705.06950}, 2017.

\bibitem{carreira2018short}
Joao Carreira, Eric Noland, Andras Banki-Horvath, Chloe Hillier, and Andrew Zisserman.
\newblock A short note about kinetics-600.
\newblock {\em arXiv preprint arXiv:1808.01340}, 2018.

\end{thebibliography}
}

\newpage
\section*{NeurIPS Paper Checklist}

\begin{enumerate}

\item {\bf Claims}
    \item[] Question: Do the main claims made in the abstract and introduction accurately reflect the paper's contributions and scope?
    \item[] Answer: \answerYes{} 
    \item[] Justification: Yes, the abstract and introduction accurately state the contribution of this work,  which is \textbf{DetectiumFire}, the largest multi-modal fire dataset.

\item {\bf Limitations}
    \item[] Question: Does the paper discuss the limitations of the work performed by the authors?
    \item[] Answer: \answerYes{} 
    \item[] Justification: The limitation of the work is presented in the Appendix~\ref{sec:discussion}

\item {\bf Theory assumptions and proofs}
    \item[] Question: For each theoretical result, does the paper provide the full set of assumptions and a complete (and correct) proof?
    \item[] Answer: \answerNA{} 
    \item[] Justification: This is a dataset paper and does not include theoretical results. 

    \item {\bf Experimental result reproducibility}
    \item[] Question: Does the paper fully disclose all the information needed to reproduce the main experimental results of the paper to the extent that it affects the main claims and/or conclusions of the paper (regardless of whether the code and data are provided or not)?
    \item[] Answer: \answerYes{} 
    \item[] Justification: All the experiment details are included in Section~\ref{sec:ex_details}. We basically follow the official implementation for each model and set most of the hyperparameters as default. The experimental result is reproducible. 

\item {\bf Open access to data and code}
    \item[] Question: Does the paper provide open access to the data and code, with sufficient instructions to faithfully reproduce the main experimental results, as described in supplemental material?
    \item[] Answer: \answerYes{} 
    \item[] Justification: We upload a sample dataset with necessary metadata of DetectiumFire in the supplemental material. 

\item {\bf Experimental setting/details}
    \item[] Question: Does the paper specify all the training and test details (e.g., data splits, hyperparameters, how they were chosen, type of optimizer, etc.) necessary to understand the results?
    \item[] Answer: \answerYes{} 
    \item[] Justification: All details can be found in Section~\ref{sec:ex_details}.

\item {\bf Experiment statistical significance}
    \item[] Question: Does the paper report error bars suitably and correctly defined or other appropriate information about the statistical significance of the experiments?
    \item[] Answer: \answerYes{} 
    \item[] Justification: We trained our object detection models using 5-fold cross validation and included the variance for each metric. 

\item {\bf Experiments compute resources}
    \item[] Question: For each experiment, does the paper provide sufficient information on the computer resources (type of compute workers, memory, time of execution) needed to reproduce the experiments?
    \item[] Answer: \answerYes{} 
    \item[] Justification: It can be found in Section~\ref{sec:ex_details}.
    
\item {\bf Code of ethics}
    \item[] Question: Does the research conducted in the paper conform, in every respect, with the NeurIPS Code of Ethics \url{https://neurips.cc/public/EthicsGuidelines}?
    \item[] Answer: \answerYes{} 
    \item[] Justification: Yes.

\item {\bf Broader impacts}
    \item[] Question: Does the paper discuss both potential positive societal impacts and negative societal impacts of the work performed?
    \item[] Answer: \answerYes{} 
    \item[] Justification: It can be found in Appendix~\ref{sec:discussion}.
    
\item {\bf Safeguards}
    \item[] Question: Does the paper describe safeguards that have been put in place for responsible release of data or models that have a high risk for misuse (e.g., pretrained language models, image generators, or scraped datasets)?
    \item[] Answer: \answerYes{} 
    \item[] Justification: It can be found in Appendix~\ref{sec:discussion}.

\item {\bf Licenses for existing assets}
    \item[] Question: Are the creators or original owners of assets (e.g., code, data, models), used in the paper, properly credited and are the license and terms of use explicitly mentioned and properly respected?
    \item[] Answer: \answerYes{} 
    \item[] Justification: It can be found in Appendix~\ref{sec:source}.

\item {\bf New assets}
    \item[] Question: Are new assets introduced in the paper well documented and is the documentation provided alongside the assets?
    \item[] Answer: \answerYes{} 
    \item[] Justification: It can be found in Appendix~\ref{sec:source}.

\item {\bf Crowdsourcing and research with human subjects}
    \item[] Question: For crowdsourcing experiments and research with human subjects, does the paper include the full text of instructions given to participants and screenshots, if applicable, as well as details about compensation (if any)? 
    \item[] Answer: \answerYes{} 
    \item[] Justification: We include details about how we curate the dataset in the main paper. 

\item {\bf Institutional review board (IRB) approvals or equivalent for research with human subjects}
    \item[] Question: Does the paper describe potential risks incurred by study participants, whether such risks were disclosed to the subjects, and whether Institutional Review Board (IRB) approvals (or an equivalent approval/review based on the requirements of your country or institution) were obtained?
    \item[] Answer: \answerNA{} 
    \item[] Justification: Not applicable. 

\item {\bf Declaration of LLM usage}
    \item[] Question: Does the paper describe the usage of LLMs if it is an important, original, or non-standard component of the core methods in this research? Note that if the LLM is used only for writing, editing, or formatting purposes and does not impact the core methodology, scientific rigorousness, or originality of the research, declaration is not required.
    \item[] Answer: \answerNA{} 
    \item[] Justification: Not applicable. 

\end{enumerate}

\newpage
\appendix

\section{Discussions}
\label{sec:discussion}
In this section, we explore potential use cases for our dataset that were not addressed in this work and highlight its limitations and broader impacts. These aspects are left as future research directions. 

\subsection{Potential Use Cases}
\paragraph{Benchmark for Synthetic Fire Data Generation}
In addition to generative techniques for synthetic fire image generation, as demonstrated in this work and by FLAME\_Diffuser~\cite{wang2024flame}, recent advancements such as Scintilla~\cite{kokosza2024scintilla} have shown promising results in simulating and rendering virtual wildfires using 3D modeling. Their approach realistically captures key aspects of the fire-burning process and its interaction with environmental factors. Experiments comparing their results to real-world wildfire data are promising. DetectiumFire can serve as a benchmark for evaluating the quality of synthetic fire generation across different methods. Furthermore, our real-world dataset can act as ground truth for comparing the realism of these synthetic outputs, offering a unique resource for benchmarking.

\paragraph{Fire Video Generation}
Building on our success with diffusion-based fire image synthesis, an exciting future direction is to extend this capability to the temporal domain using video diffusion models. By learning spatio-temporal patterns from fire video clips, these models could generate realistic, temporally coherent sequences that simulate fire behavior over time. Such generated fire videos would be immensely valuable for applications in safety simulation, risk prediction, and synthetic training of surveillance or anomaly detection systems, particularly in domains where real-world dangerous fire footage is scarce or ethically infeasible to capture. Furthermore, conditioning the generation process on control signals, such as fire type, spread rate, environmental conditions, or suppression methods, could unlock customizable simulations that serve not only computer vision research, but also training for first responders, fire risk modeling, and digital twin systems in smart infrastructure~\cite{khajavi2023digital}.

\paragraph{Advanced Fire Reasoning with Vision-Language Models and AI Agents}
While our current work demonstrates the feasibility of training vision-language models (VLMs) to infer basic fire attributes such as burning object, environment, and severity, future directions involve integrating these models into more advanced AI agents capable of dynamic and autonomous fire scene understanding. Such agents would not only interpret static images or single frames but also perform \textbf{temporal reasoning}, predicting how a fire is likely to progress over time. They could engage in \textbf{causal inference}, identifying ignition sources or environmental factors accelerating fire spread. When combined with external knowledge bases and decision logic, these agents could offer real-time decision support, advising on context-aware actions, such as whether to trigger alarms, evacuate, or suppress a fire based on observed conditions.

Moreover, fire-aware AI agents could interact with humans through multi-turn dialogues, generate structured incident reports, or simulate different intervention outcomes under uncertainty. These capabilities would transform passive VLMs into proactive, embodied reasoning systems capable of operating in safety-critical domains like smart buildings, industrial control rooms, or search-and-rescue missions. Enabling such agents requires richer datasets like DetectiumFire, augmented with temporal context, multi-modal annotations, and high-level task specifications, laying the foundation for the next generation of fire-resilient intelligent systems.

\subsection{Limitations and Broader Impacts}
Despite our efforts to ensure the quality, safety, and diversity of DetectiumFire, several inherent limitations remain. As described in Section~\ref{sec:real}, our data pipeline includes quality analysis, filtering, and human verification. However, due to the inherently extreme and unpredictable nature of fire scenes, some images and captions may still contain content that is unsafe, misleading, or potentially harmful. This introduces a risk that models trained on the dataset could learn to generate unsafe or inappropriate outputs, raising concerns about alignment and responsible deployment, especially in generative settings.

Another limitation arises from data sourcing. A large portion of our real-world dataset is collected via web search, which may impose legal or ethical restrictions on downstream usage. Although we have made every effort to ensure compliance with licensing and proper attribution, this limitation may constrain the dataset's applicability to strictly academic or non-commercial research.

Furthermore, while DetectiumFire covers a broad spectrum of fire types and scenes, it is still not exhaustive. Edge cases, underrepresented geographies, and culturally specific fire-related scenarios may be missing or under-sampled.

Nonetheless, we believe that releasing DetectiumFire as an open dataset offers substantial positive impact. With recent wildfires and industrial fires causing widespread devastation, there is an urgent need for AI systems that can understand, detect, and respond to fire events in a reliable and context-aware manner. We hope that DetectiumFire can support this mission by enabling the research community to develop and fine-tune models with a deeper, more nuanced understanding of fire dynamics, ultimately contributing to safer AI deployments in smart homes, infrastructure monitoring, public safety, and environmental protection.

\section{Additional Details of DetectiumFire}

\subsection{DetectiumFire Real-World Data Sources, Liscening and Ethical Considerations}
\label{sec:source}
Table~\ref{tab:data_source} shows a detailed breakdown of all data sources, including licensing information and any modifications performed on each dataset. We again fully acknowledge the contributions of prior datasets and ensure adherence to all stated licenses. 

\begin{table}[ht]
\centering
\caption{Summary of real-world data sources in DetectiumFire.}
\label{tab:data_source}
\renewcommand{\arraystretch}{1.2}
\setlength{\tabcolsep}{4pt}
\begin{tabular}{@{}p{2.8cm} p{3cm} p{2.3cm} p{5cm}@{}}
\toprule
\textbf{Data Source} & \textbf{Size (Img/Vid)} & \textbf{License} & \textbf{Notes} \\
\midrule
Web Search & 12,920 / 2,468 & CC BY-NC-SA 4.0 & Collected from Google, YouTube, TikTok, Twitter \\
IoT Device & 318 / 0 & CC BY-NC-SA 4.0 & Captured using IoT sensors in controlled demos \\
FIRE Dataset\footnotemark[1] & 433 / 0 & CC0 & Originally for classification; we added bboxes + prompts \\
Forest Fire\footnotemark[2] & 373 / 0 & CC BY 4.0 & Originally classification; we added bboxes + prompts \\
FireNET\footnotemark[3] & 485 / 0 & MIT & Originally object detection; bboxes modified + prompts added \\
\bottomrule
\end{tabular}
\end{table}

\footnotetext[1]{\url{https://www.kaggle.com/datasets/phylake1337/fire-dataset}}
\footnotetext[2]{\url{https://www.kaggle.com/datasets/alik05/forest-fire-dataset}}
\footnotetext[3]{\url{https://github.com/OlafenwaMoses/FireNET}}

Notice that most part of our real-world data was collected via public web search. In accordance with standard academic practice and ethical use guidelines, these data should be restricted to \textbf{academic use only}.

\subsection{Annotation Quality and Inter-Annotator Agreement}
\label{sec:anno_quality}

Our annotation pipeline is structured around three key tasks:

\begin{enumerate}
    \item Bounding box annotation of fire regions in the images.
    \item Caption generation: Creating 75-token descriptive captions covering what is burning, where it is burning, and its severity (no/low/moderate/high risk).
    \item Preference annotation: Selecting preferred images between pairs of synthetic samples generated by different diffusion models and justifying the choice based on general preference, visual appeal, and prompt alignment.
\end{enumerate}

All annotators are affiliated with Detectium~\footnote{\url{https://www.detectium.io/}}, a startup focused on reducing industrial fire-related risks through digital twinning and AI technologies. Annotators either possess prior fire-related domain experience or have participated in related projects and are familiar with basic computer vision concepts.

During annotation, each annotator will be allocated one batch of data. To ensure quality, any ambiguities or disagreements during annotation are recorded and discussed with the technical officers. Once an annotator completes a batch of data, it is automatically routed for review. Each annotation is checked, and a minimum 90\% agreement rate is required for batch approval. This includes bounding box accuracy, caption accuracy (coverage of key fire scene elements), and preference validity based on the established criteria (general preference, visual appeal, and prompt alignment). We chose the 90\% threshold acknowledging that 100\% agreement is neither practical nor necessarily desirable, as it may introduce reviewer bias. If a batch falls short of this threshold, the technical officer and the annotator jointly discuss the issues, and the data is re-annotated.

\subsection{GPT Prompts for Captioning}
\label{sec:caption}
While it is certainly possible to write all captions manually, the cost-to-benefit ratio is not favorable in our current setup. Given the strong performance of modern GPT models, often approaching human-level quality, and their widespread adoption in research for automating annotation tasks~\cite{xu2024llavacot,hou2024assessing,tan2024large}, we leverage them to reduce manual workload. Specifically, GPT-generated captions, when followed by careful human review and editing, offer an effective balance between quality and scalability. To this end, we first leverage GPT-4o~\cite{openai2023gpt4} to generate detailed and semantically rich descriptions. Notice that we also instruct the model to infer contextual information such as the presence of people in our prompt; however, this information is excluded from the current dataset and reserved for future work involving advanced reasoning tasks. The full prompt text used for this process is provided below.

\textbf{System Instruction:} You are an expert in fire scene analysis. Given an image, analyze its content and return a structured JSON response.

\vspace{0.5em}
\textbf{Instructions:}
\begin{enumerate}
    \item \textbf{Provide a detailed description} of the image.
    \item \textbf{If the image does not contain fire}, return only the description.
    \item \textbf{If the image contains fire}, answer the following questions in a structured format:
    \begin{itemize}
        \item \textbf{Objects on Fire}: Identify the specific objects or materials that are burning.
        \item \textbf{Fire Severity Level}: Classify the fire into one of the following categories:
        \begin{itemize}
            \item \texttt{Controlled Fire (No Risk)}: A small fire used for cooking, lighting, or heating, posing no immediate danger.
            \item \texttt{Minor Fire (Low Risk)}: A small, contained fire (e.g., small trash fire) that could be extinguished easily.
            \item \texttt{Moderate Fire (Medium Risk)}: A fire spreading but still manageable, posing some risk to nearby objects or people.
            \item \texttt{Severe Fire (High Risk)}: A large, uncontrolled fire that is rapidly spreading and endangering structures or lives.
        \end{itemize}
        \item \textbf{Affected Area}: Describe the extent of the fire’s impact (e.g., a single object, a room, an entire building).
        \item \textbf{Presence of People}: Indicate whether people are visible in the scene and describe their condition (e.g., evacuating, trapped, firefighting).
    \end{itemize}
\end{enumerate}

\vspace{1em}
\textbf{Expected JSON Output Format:}

\vspace{0.5em}
\textbf{If fire is present:}
\begin{verbatim}
{
  "description": "A large fire engulfing a residential building with thick black smoke.",
  "objects_on_fire": "House, furniture",
  "fire_severity": "Severe Fire (High Risk)",
  "affected_area": "The entire house and nearby trees",
  "people_present": "Yes, firefighters are attempting to control the fire."
}
\end{verbatim}

\vspace{0.5em}
\textbf{If no fire is present:}
\begin{verbatim}
{
  "description": "A quiet urban street with parked cars and people walking on the sidewalk."
}
\end{verbatim}

\vspace{0.5em}
\textbf{Always ensure the response follows the correct JSON format.}

\vspace{0.5em}
\noindent \textbf{[Image]} \\
\texttt{{image}}

\vspace{0.5em}
\noindent \textbf{[Answer]}

\subsection{Annotation Tool}
\label{sec:anno_tool}
\begin{figure}[ht]
    \centering
    \includegraphics[width=1\linewidth]{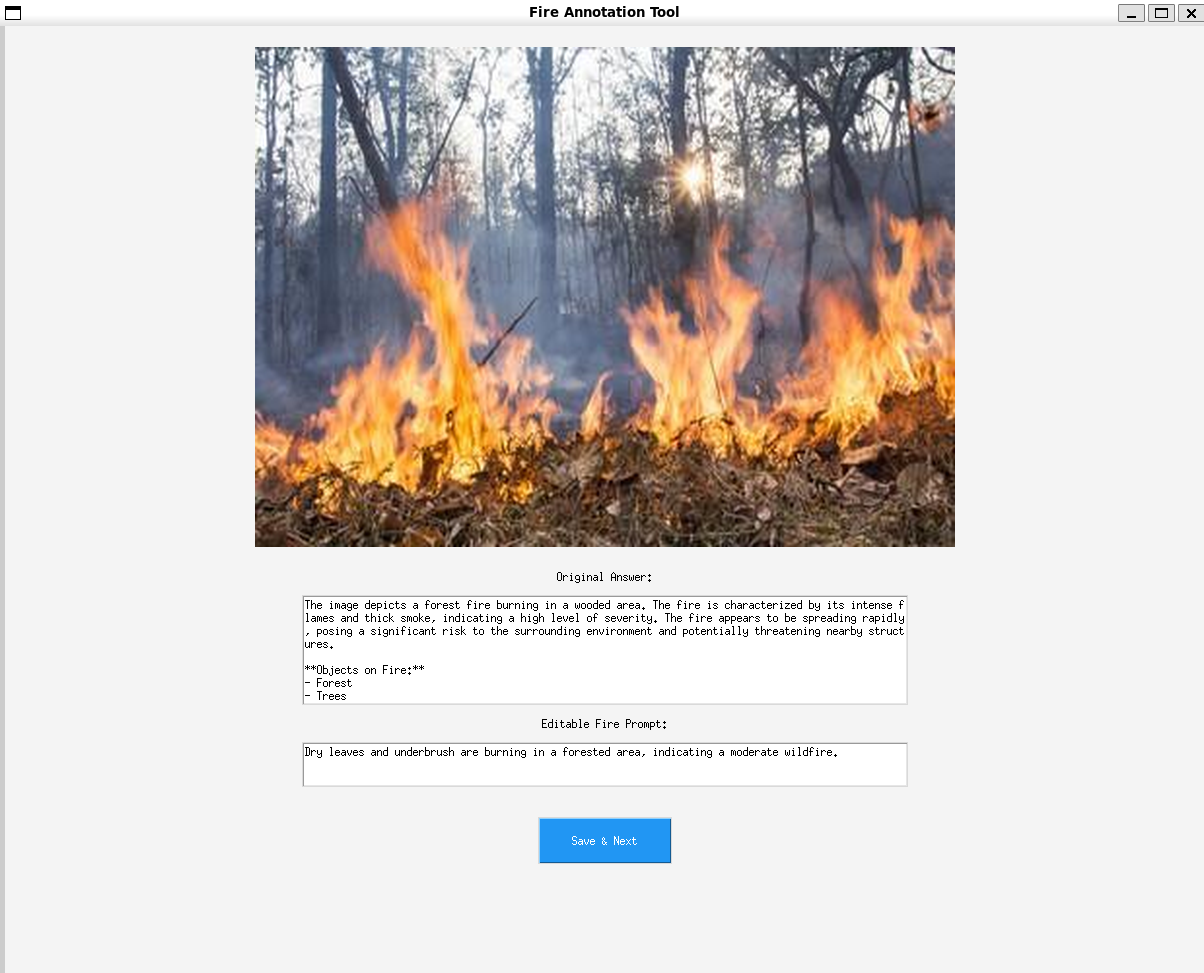}
    \caption{Screenshot of the fire annotation tool. The tool displays each image alongside the original GPT-generated caption and allows human annotators to edit and finalize a concise, domain-specific prompt.}
    \label{fig:annotation_tool}
\end{figure}

To facilitate efficient and high-quality prompt annotation, we developed a lightweight Python-based GUI application for human verification and editing (see Figure~\ref{fig:annotation_tool}). The tool displays each image alongside two text boxes: the first presents the original caption generated by GPT-4o, including detailed fire-related descriptions such as burning objects and severity level; the second allows human annotators to review and edit the caption to produce a final, concise fire-specific prompt. 

\subsection{Fire Scene Taxonomy in DetectiumFire}
\label{sec:tax}

We divide scenes into two major categories: Indoor Fires (3,374 images) and Outdoor Fires (4,175 images), and further organize them by ignition source and situational relevance. The detail breakdown can be found in Table~\ref{tab:fire_categories}.

\begin{table}[ht]
\centering
\caption{Detailed taxonomy of real-world fire scenarios in DetectiumFire.}
\label{tab:fire_categories}
\small
\renewcommand{\arraystretch}{1.2}
\begin{tabularx}{\textwidth}{@{}lXr@{}}
\toprule
\multicolumn{3}{l}{\textbf{Indoor Fires (Total: 3,374 images)}} \\
\midrule
\textbf{Subcategory} & \textbf{Description} & \textbf{Image Count} \\
\midrule
Cooking fire (controlled) & Stove-top flames from regular cooking & 54 \\
Stove/fireplace fire & Contained stove or heater fires & 30 \\
Candle flame & Lit candles in normal use & 1,076 \\
Lighter flame & Small flame from lighters & 1,075 \\
Matches flame & Small ignition flames from matchsticks & 163 \\
Kitchen fire (uncontrolled) & Fires spreading from cookers or frying oil & 212 \\
Electrical fire & Fires from phones, fans, wire, etc. & 213 \\
Other indoor fires (accidental) & Dropped candles, unknown cause fires & 551 \\
\midrule
\multicolumn{3}{l}{\textbf{Outdoor Fires (Total: 4,175 images)}} \\
\midrule
Campfire (controlled) & Normal campfires outdoors & 347 \\
Vehicle fires & Fires from cars, trucks, motorcycles & 750 \\
Forest/wildfires & Large outdoor or forest-based fires & 1,045 \\
House/residential fires & Full or partial house fires & 1,159 \\
Ship fires & Fire incidents on marine vessels & 42 \\
Plane fires & Fires on aircraft (in-air or grounded) & 14 \\
Trash bin fires & Fires in dumpsters or public trash bins & 26 \\
LPG/gas tank fires & Fires from liquefied gas canisters (indoor/outdoor) & 159 \\
Burning debris & Burning of leaves, paper, woodpiles in the open & 190 \\
Other outdoor fires & Scenes not covered by the above & 418 \\
Pure flame only & Cropped flames without scene context & 25 \\
\bottomrule
\end{tabularx}
\end{table}

\subsection{Additional Examples from DetectiumFire}
\label{sec:add_example}
\begin{figure}[ht]
    \centering
    \begin{subfigure}[b]{0.45\textwidth}
        \includegraphics[width=\textwidth]{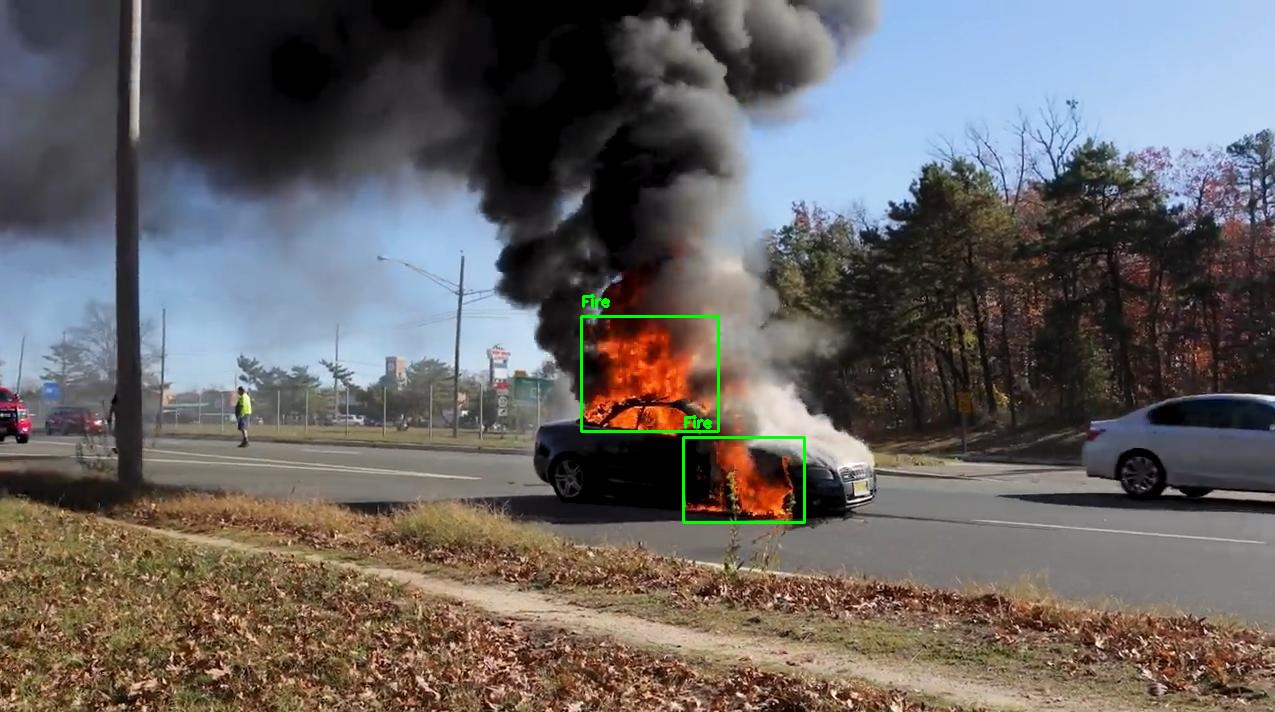}
        \caption{A vehicle is fully engulfed in flames on a public road, emitting thick black smoke. The fire is high risk, severe and uncontrolled.}
    \end{subfigure}
    \hfill
    \begin{subfigure}[b]{0.45\textwidth}
        \includegraphics[width=\textwidth]{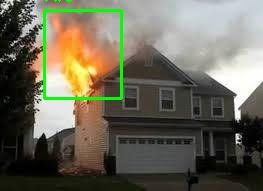}
        \caption{Flames are erupting from the upper portion of a residential building, smoke is spreading into the sky, indicating high fire risk.}
    \end{subfigure}

    \vspace{0.5em}

    \begin{subfigure}[b]{0.45\textwidth}
        \includegraphics[width=\textwidth]{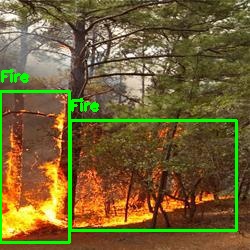}
        \caption{A wildfire is actively burning in a forested area, with visible flames spreading across dry underbrush and tree trunks. It is a high-risk outdoor fire scenario.}
    \end{subfigure}
    \hfill
    \begin{subfigure}[b]{0.45\textwidth}
        \includegraphics[width=\textwidth]{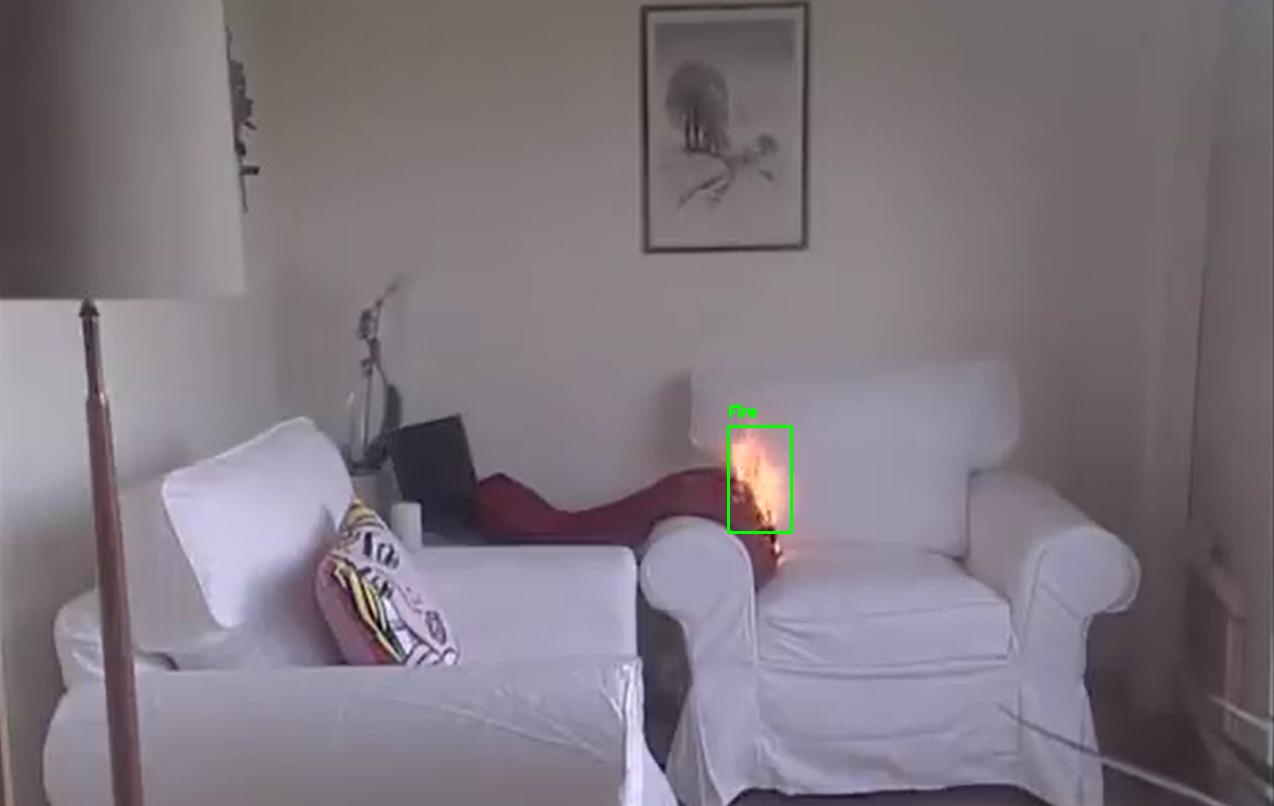}
        \caption{A small flame is observed on a couch cushion in a living room. The fire currently poses moderate risk.}
    \end{subfigure}

    \caption{Representative samples from DetectiumFire showcasing diverse fire scenarios across both indoor and outdoor settings.}
    \label{fig:fire_examples}
\end{figure}

To visually illustrate the breadth and realism of fire scenarios captured in DetectiumFire, we showcase four representative samples from our dataset in Figure~\ref{fig:fire_examples}. These examples span a range of indoor and outdoor fire contexts, with high-quality bounding box annotations and captions. These diverse, real-world examples highlight the fine-grained coverage and detailed annotation quality of DetectiumFire, reinforcing its utility for both traditional computer vision tasks and advanced reasoning in fire-related safety applications.

\begin{figure}[ht]
    \centering
    \includegraphics[width=0.8\linewidth]{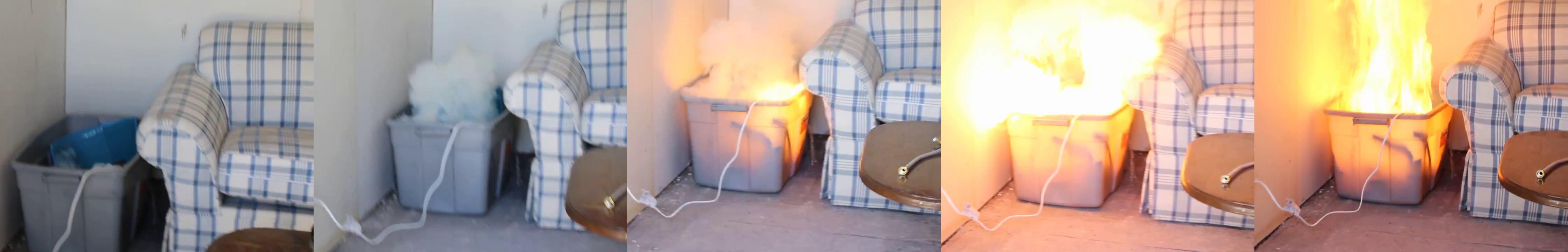}
    
    \includegraphics[width=0.8\textwidth]{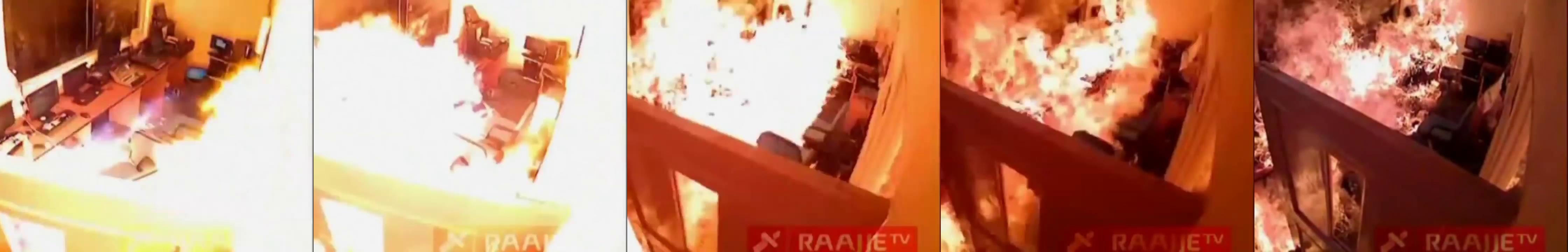} \\ 
    
    \includegraphics[width=0.8\textwidth]{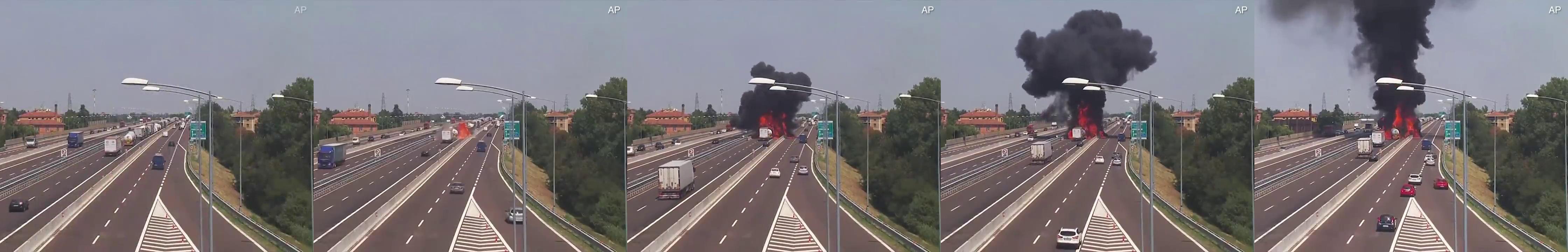} \\ 
    
    \includegraphics[width=0.8\textwidth]{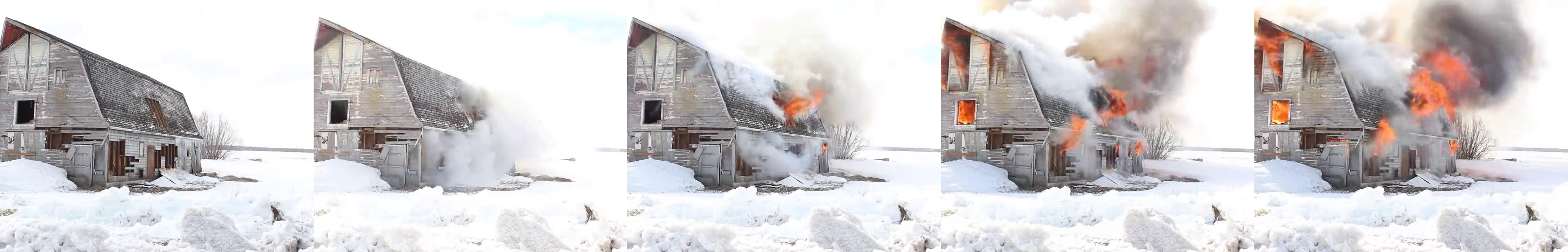} \\ 
    \caption{Additional example videos from the DetectiumFire dataset showcase various fire scenes. }
    \label{fig:video}
\end{figure}

Figure~\ref{fig:video} shows additional example videos from the DetectiumFire dataset. The reason for collecting video for fire examples is that, in many cases, a single image is insufficient for accurately recognizing fire. For instance, at the early stages of fire ignition, as seen in the video clips in Figure~\ref{fig:video}, distinguishing fire can be challenging. Similarly, in low-light environments or with low-quality cameras, bright light sources can appear indistinguishable from fire, as demonstrated in the second video. However, fire exhibits unique dynamic motion patterns in videos, such as flowing or flickering, whereas other objects like light sources tend to remain static. This makes the video dataset a critical complement to the image dataset for capturing the temporal characteristics of fire, which are essential for a deeper understanding of fire dynamics.

\begin{figure}[ht]
    \centering
    \begin{subfigure}[b]{0.23\textwidth}
        \includegraphics[width=\textwidth]{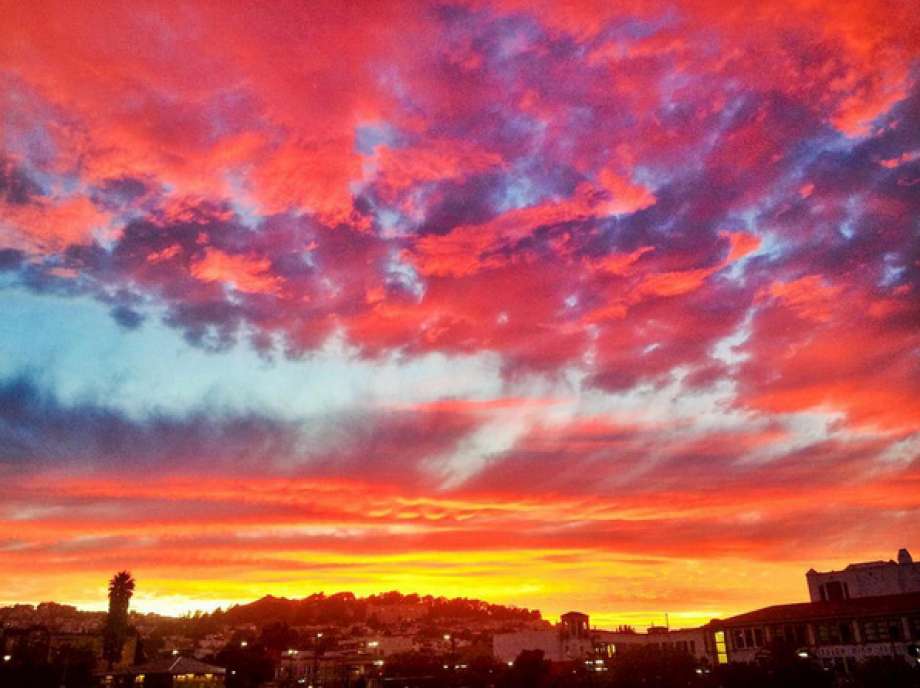}
        \caption{Sunset sky}
    \end{subfigure}
    \begin{subfigure}[b]{0.23\textwidth}
        \includegraphics[width=\textwidth]{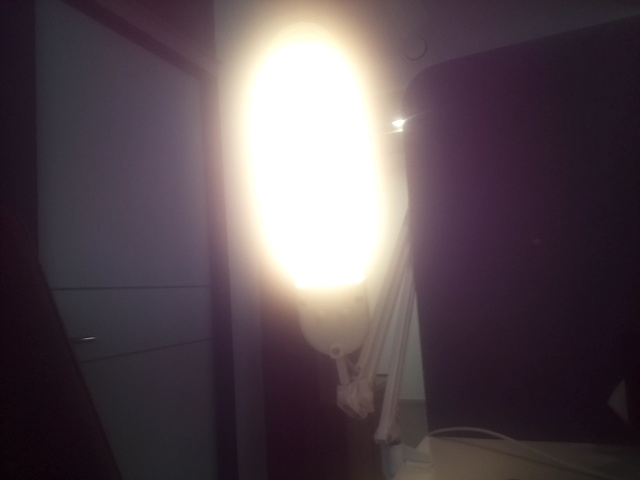}
        \caption{Light in the dark}
    \end{subfigure}
    \begin{subfigure}[b]{0.26\textwidth}
        \includegraphics[width=\textwidth]{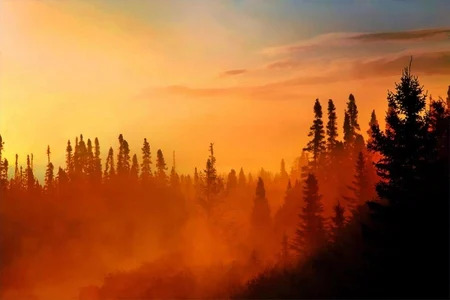}
        \caption{Red mist in the forest}
    \end{subfigure}
    \begin{subfigure}[b]{0.23\textwidth}
        \includegraphics[width=\textwidth]{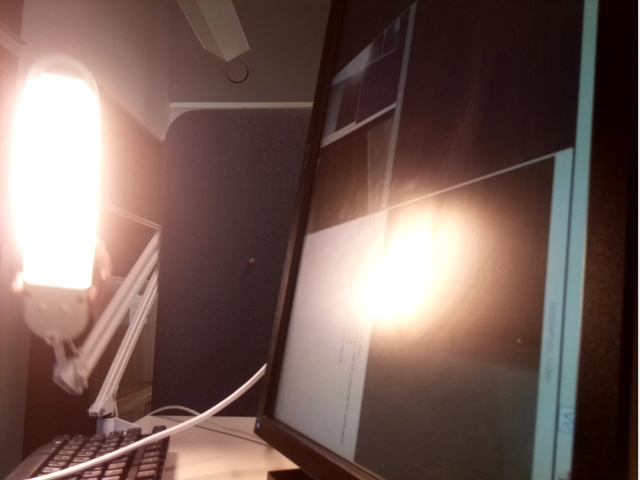}
        \caption{Red reflection}
    \end{subfigure}
    \caption{Examples of non-fire images that commonly trigger false positives in current fire detection models.}
    \label{fig:false_positives}
\end{figure}

In addition to collecting diverse fire scenes, DetectiumFire also includes a curated set of non-fire images specifically selected to address common sources of false positives in existing fire detection models. Unlike conventional negative samples, which may include arbitrary non-fire scenes, we deliberately incorporate visually ambiguous cases that often confuse models due to their fire-like visual characteristics. As shown in Figure~\ref{fig:false_positives}, such examples include red-hued skies during sunset, intense light sources in dark environments, fog or mist with warm color tones, and reflective surfaces that mimic the glow of fire. By including these hard negatives, we aim to improve model robustness and reduce false alarms in real-world deployments, particularly in safety-critical settings where false positives can lead to costly outcomes.

\section{Additional Experiment Setup}
\subsection{Calculating Elo Ranking}
\label{sec:elo}

The Elo rating system, originally developed for chess, provides a straightforward yet effective framework for estimating relative skill levels in head-to-head competitions~\cite{khan2024debating}. The algorithm assumes that the performance of each competitor can be modeled as a normally distributed random variable. Then the expected outcome of a match is calculated as a logistic function of the difference between the ratings of the two players. Specifically, define the Elo ratings for one competitor $C_1$ as $E_1$ and $E_2$ for the other competitor $C_2$, then the probability of $C_1$ winning against $C_2$ is computed as:
\begin{equation}
    \tilde{\omega}_{C_1}=\frac{1}{1+10^{\frac{E_2-E_1}{400}}}
\end{equation}

The formula produces a value between 0 and 1, representing the likelihood of $C_1$ defeating $C_2$, which is $\tilde{\omega}_{C_1}$. 

To compute the optimal Elo ratings, one simple method is to update the scores based on the outcome after each comparison. However, to better refine Elo ratings over a historical dataset, we employ an optimization approach to best fit the ratings to observed outcomes. The goal of the optimizing process involves minimizing the difference between the predicted and observed win rates in the dataset. We define the loss as a square error between the predicted win rate $\tilde{\omega}$, derived from Elo ratings, and the actual win rate, $\omega$, observed in the dataset:
\begin{equation}
    L(\tilde{\omega}) = \sum_{(C_i, C_j)} (\tilde{\omega}_{C_i}-\omega_{C_i, C_j})^2
\end{equation}

where $\tilde{\omega}_{C_i}$ is the predicted win rate of competitor $C_i$, $\omega_{C_i, C_j}$ is the actual win rate of competitor $C_i$ against competitor $C_j$ in the dataset. 

In this work, we use the Broyden-Fletcher-Goldfarb-Shanno (BFGS) algorithm. We first initialize the ratings for all competitors at a default value of 1200. Then we iteratively refine the ratings by minimizing the loss function. To give an insight into the uncertainty and robustness of the Elo ratings, confidence intervals are estimated using statistical bootstrapping by resampling different subsets of matchups and recalculating the Elo ratings for 1000 iterations.

\subsection{GPT Prompts for Evaluation}
\label{sec:gpt_prompt}
To evaluate the alignment of the three criteria (General Preferences, Visual Appeal, and Prompt Alignment) with actual human preferences, we compared the original Stable Diffusion v1.5 model against versions fine-tuned using Supervised Fine-Tuning (SFT) and Reinforcement Learning from Human Feedback (RLHF). Given the impracticality of relying solely on human evaluators for large-scale evaluations, we utilized GPT-4o~\cite{openai2023gpt4} as an automated judge. For consistency and reproducibility, we designed detailed GPT-4o prompts to evaluate the generative performance of the models. The full text of these prompts is provided below:

\textbf{User Prompt:}

\textbf{[System]}

You are tasked with evaluating two images based on the following three criteria, with a specific focus on the \textbf{quality of fire generation}. Ignore other elements in the image (e.g., humans or unrelated background objects) and assess solely based on the quality, realism, and prompt alignment of the fire depiction.

1. \textbf{General Preference}: How visually appealing and convincing the generated fire appears overall.

2. \textbf{Visual Appeal}: The artistic quality, realism, and aesthetic beauty of the generated fire.

3. \textbf{Prompt Alignment}: How accurately the generated fire matches the intent and details described in the provided prompt.

\textbf{Instructions}:

\begin{itemize}
  \item First, output the scores (between 0--10) for each image. Use the exact format:
  
  \texttt{[[score11, score12, score13], [score21, score22, score23]]}
  
  \begin{itemize}
    \item \texttt{score11}: First criteria score for Image 1
    \item \texttt{score12}: Second criteria score for Image 1
    \item \texttt{score13}: Third criteria score for Image 1
    \item \texttt{score21}: First criteria score for Image 2
    \item \texttt{score22}: Second criteria score for Image 2
    \item \texttt{score23}: Third criteria score for Image 2
  \end{itemize}
  
  \item Then, provide a \textbf{short explanation} comparing the two images, focusing only on the fire generation quality.

  \item Avoid any order biases, and ensure the order in which the images are presented does not influence your judgment.
\end{itemize}

\textbf{Example Output:} \\
\texttt{[[8, 9, 7], [7, 8, 9]]} 

The first image features a more vibrant and realistic fire with clear flames and smoke, making it visually appealing and convincing overall. The second image, while still visually appealing, lacks the same level of detail in the flames and smoke, and its alignment with the prompt is slightly less accurate.

\textbf{Important Note:}
\begin{itemize}
  \item Focus exclusively on the \textbf{fire generation quality} when evaluating.
  \item Give under-5 scores for pictures without fire.
  \item Ignore unrelated elements such as humans, background objects, or other non-fire elements.
\end{itemize}

\vspace{1em}
\noindent \textbf{[Prompt]} \\
\texttt{{prompt}}

\vspace{0.5em}
\noindent \textbf{[Image 1]} \\
\texttt{{image 1}}

\vspace{0.5em}
\noindent \textbf{[Image 2]} \\
\texttt{{image 2}}

\vspace{0.5em}
\noindent \textbf{[Answer]}

\section{Additional Experiment Results}

\subsection{Video Understanding Results}
\label{sec:video_ex}
This experiment demonstrates the applicability of DetectiumFire for anomaly detection, where fire is commonly treated as an anomalous event in surveillance and industrial contexts. Given the lack of established baselines for fire-related anomaly detection, especially using multi-modal data, we provide foundational results using two standard video classification models: TimeSformer~\cite{gberta_2021_ICML} and VideoMamba~\cite{li2024videomamba}. TimeSformer is a state-of-the-art framework based on space-time attention and has shown strong performance on anomaly detection datasets like Kinetics-400~\cite{kay2017kinetics}, Kinetics-600~\cite{carreira2018short}. VideoMamba is a structured state space model (SSM)-based video backbone designed as an efficient alternative to transformers for video understanding. While our current setup avoids complex temporal reasoning, it serves to isolate dataset usability under basic configurations. Future work should explore more sophisticated spatio-temporal models for deeper insights. 

Specifically, we train three TimeSformer variants using divided space-time attention, each operating on clips with 8, 16, or 96 frames, and with input resolutions of 224×224, 448×448, and 224×224 pixels, respectively. In addition, we fine-tune three variants of VideoMamba: Ti, S, and M with a batch size of 4 and vary the number of frames (8, 16). All other hyperparameters follow the default settings. The video data is split into 70\% training, 20\% validation, and 10\% test sets. Model selection is based on validation loss, and final results are reported on the test set.

Table~\ref{tab:timesformer} shows the classification accuracy of the three TimeSformer variants. Since the task involves a single target class (fire), it is relatively straightforward, and all three models achieve consistently high accuracy. The base model, which operates with minimal computational complexity by using only 8 frames and a spatial crop size of 224, achieves an accuracy of 96.6\%. As we increase the temporal resolution (number of frames) and the spatial crop size, we observe a consistent improvement in performance. The TimeSformer-HR model, which doubles the temporal resolution to 16 frames and increases the spatial crop size to 448, achieves a 0.8\% improvement, resulting in an accuracy of 97.4\%. The TimeSformer-L model, which significantly increases the temporal resolution to 96 frames while maintaining the baseline spatial crop size of 224, achieves the highest accuracy of 97.9\%.

Table~\ref{tab:videomamba} shows the classification accuracy of several VideoMamba variants. Interestingly, while VideoMamba demonstrates solid performance across different configurations, we observe that it consistently underperforms TimeSformer on our fire dataset, which is not expected given its strong performance on general-purpose video benchmarks. This suggests that VideoMamba may not transfer as effectively to the fire understanding domain. Some potential reasons for this could be: 1) VideoMamba’s state space design excels at modeling smooth and continuous temporal dynamics, which are prevalent in action recognition datasets like Kinetics~\cite{kay2017kinetics}. However, fire-related events, such as sudden flare-ups, flickering, or rapid spreading, often involve chaotic and non-stationary motion patterns. These may be better captured by transformer-based models that attend flexibly across time. 2) The Mamba architecture is known for its efficiency due to low-rank approximations and memory compression. While this helps scale to long sequences, it may inadvertently discard fine-grained temporal details crucial for fire scene interpretation, such as frame-level intensity shifts or flickering textures.

In summary, these findings highlight the importance of tailoring temporal modeling architectures to the unique characteristics of fire-related video data. Future work may explore hybrid models that combine the efficiency of Mamba with the flexibility of temporal attention techniques to better bridge the gap between general video datasets and fire-specific understanding.

\begin{table}[ht]
\caption{Performance of different TimeSformer models with various hyperparameters on our video dataset.}
\vskip 0.15in
    \centering
    
    \begin{small}
    \begin{tabular}{lcccc}
        \toprule
        \textbf{Method} & \textbf{\# of Frames} & \textbf{Spatial Crop} & \textbf{Acc.}  \\
        \midrule
        TimeSformer & 8&  224& 96.6\\
        
        TimeSformer-HR & 16 &   448& 97.4\\

        TimeSformer-L & 96 & 224 & 97.9 \\
        
        \bottomrule
    \end{tabular}
    \end{small}
    
    \label{tab:timesformer}
    \vskip -0.1in
\end{table}

\begin{table}[ht]
\caption{Performance of different VideoMamba models with various hyperparameters on our video dataset.}
\vskip 0.15in
    \centering
    
    \begin{small}
    \begin{tabular}{lcccc}
        \toprule
        \textbf{Method} & \textbf{Pretraining} & \textbf{Resolution}  & \textbf{\# of Frames}  & \textbf{Acc.}  \\
        \midrule
        VideoMamba-Ti & ImageNet-1k&  224& 8$\times$3$\times$4 &81.35 \\
        
         VideoMamba-Ti & ImageNet-1k&  224& 16$\times$3$\times$4 &83.13 \\

         VideoMamba-S & ImageNet-1k&  224& 8$\times$3$\times$4 &84.93 \\
         VideoMamba-S & ImageNet-1k&  224& 16$\times$3$\times$4 &85.27 \\
        
         VideoMamba-M & ImageNet-1k&  224& 8$\times$3$\times$4 &85.16 \\

         VideoMamba-M & ImageNet-1k&  224& 16$\times$3$\times$4 &86.04 \\
        
        \bottomrule
    \end{tabular}
    \end{small}
    
    \label{tab:videomamba}
    \vskip -0.1in
\end{table}

\subsection{Understanding YOLOv11 vs. Faster R-CNN Performance Disparity}\label{sec:yolo_fast}
Table~\ref{tab:object} shows that YOLOv11 underperforming Faster R-CNN in both the real-only and real+synthetic settings. We believe this is due to the unique nature of fire detection, where the target objects are often non-rigid, partially occluded, and visually ambiguous. Faster R-CNN, as a two-stage detector with explicit region proposals, tends to be more robust in such cases, particularly on medium-scale datasets with high variation and subtle object boundaries. In contrast, YOLOv11, while a state-of-the-art one-stage model, may struggle with diffuse fire regions that lack clearly defined edges or shapes. 

Moreover, our dataset includes a broad range of fire scales and environments (from small indoor flames to large outdoor fires), and this variability may favor the more structured, proposal-based architecture of Faster R-CNN in certain metrics. 

This observation also highlights that raw model capacity is not always the best indicator of performance in safety-critical, domain-specific tasks.

\subsection{FLAME\_SD vs DetectiumFire Performance}
\label{sec:flame_detectium}

Table~\ref{tab:object} shows that the Flame\_SD dataset~\cite{wang2024flame} yields significantly worse results than all other datasets. We believe the performance gap highlights important differences in dataset generation strategy, data quality, and coverage.

FLAME\_SD constructs fire masks using simple geometric shapes (e.g., rectangles, circles), without leveraging any semantic understanding of the background image. As a result, fire is often generated in unrealistic or contextually irrelevant locations, which limits its utility for training object detectors. Moreover, fire lacks fixed structure, and constraining it to predefined mask shapes weakens the expressiveness of the diffusion model,  contributing further to poor generalization.

Additionally, FLAME\_SD focuses primarily on outdoor and large-scale fire scenarios such as forest fires, while lacking diversity in small fires, indoor scenes, industrial contexts, or subtle object-fire interactions. In contrast, DetectiumFire includes a broader range of fire types and environments, leading to more balanced and effective coverage of real-world fire understanding scenarios. This discrepancy in coverage likely explains why models trained on FLAME\_SD generalize poorly when evaluated on DetectiumFire.

\subsection{Elo Results}
\label{sec:elo_result}
\begin{table*}[ht]
\centering
\caption{Pairwise Elo scores and confidence intervals for evaluating synthetic fire image generation across three criteria: General Preference, Visual Appeal, and Prompt Alignment.}
\label{tab:elo_scores}
\renewcommand{\arraystretch}{1.3}
\resizebox{\textwidth}{!}{
\begin{tabular}{@{}llccc@{}}
\toprule
\textbf{Model Pair} & \textbf{Criterion} & \textbf{Model A (Score, CI)} & \textbf{Model B (Score, CI)} \\
\midrule
\multirow{3}{*}{SFT vs DPO} 
& General Preference & 1161.99 (1148.76, 1225.18) & 1238.01 (1174.82, 1251.24) \\
& Visual Appeal & 1157.56 (1144.92, 1228.78) & 1242.44 (1171.21, 1255.08) \\
& Prompt Alignment & 1145.17 (1133.04, 1242.68) & 1254.83 (1157.32, 1266.96) \\
\midrule
\multirow{3}{*}{SFT vs Original} 
& General Preference & 1205.49 (1192.16, 1218.82) & 1194.51 (1181.19, 1193.83) \\
& Visual Appeal & 1210.30 (1201.60, 1222.06) & 1189.71 (1177.95, 1198.39) \\
& Prompt Alignment & 1213.05 (1198.39, 1224.85) & 1186.95 (1175.15, 1201.60) \\
\midrule
\multirow{3}{*}{DPO vs Original} 
& General Preference & 1212.39 (1200.46, 1225.37) & 1187.61 (1175.69, 1199.53) \\
& Visual Appeal & 1214.50 (1197.91, 1227.02) & 1185.50 (1172.97, 1202.09) \\
& Prompt Alignment & 1206.54 (1194.82, 1218.26) & 1193.46 (1181.74, 1203.24) \\
\bottomrule
\end{tabular}
}
\end{table*}

Table~\ref{tab:elo_scores} presents the pairwise Elo scores with confidence intervals across three key evaluation criteria: General Preference, Visual Appeal, and Prompt Alignment, for synthetic fire images generated by models fine-tuned via Supervised Fine-Tuning (SFT) and Reinforcement Learning from Human Feedback (RLHF/DPO), as well as the original Stable Diffusion model.

Overall, the results show that both SFT and DPO consistently outperform the original model across all criteria, confirming that fine-tuning with DetectiumFire improves generation quality. Between the two fine-tuning methods, DPO exhibits a slight but consistent advantage over SFT, particularly in Prompt Alignment and Visual Appeal. For instance, in the SFT vs DPO comparison, DPO achieves higher Elo scores across all criteria, with a noticeable margin in Prompt Alignment (1254.83 vs. 1145.17).

Interestingly, while the differences between SFT and the original model are relatively modest, the DPO vs Original comparison shows the most pronounced gains, suggesting that RLHF offers stronger alignment with user preferences for fire scene generation. These trends demonstrate the effectiveness of human feedback in improving both fidelity and prompt consistency in generative fire modeling.

\subsection{Image Comparison Examples rated by GPT-4o}
\label{sec:compare}
Below, we provide detailed examples of scoring and justification from GPT-4o.  

\newpage
\begin{table}[h!]
\caption{Comparison of Stable Diffusion 1.5 fine-tuned using SFT and RLHF. Ratings and justification are based on GPT-4o evaluation.}
\vskip 0.15in
\centering
\renewcommand{\arraystretch}{1.5} 
\setlength{\tabcolsep}{10pt} 

\begin{tabular}{|p{15cm}|}
\hline
\multicolumn{1}{|c|}{\textbf{Text Prompt:} Trees are burning next to some brush} \\ \hline

\textbf{Image 1: Stable Diffusion 1.5 (SFT)} \\ 
\makebox[\textwidth]{\includegraphics[width=0.4\textwidth]{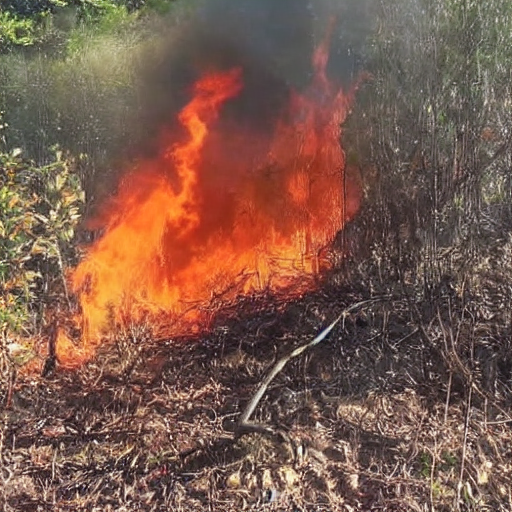}} \\ \hline

\textbf{Image 2: Stable Diffusion 1.5 + RLHF} \\
\makebox[\textwidth]{\includegraphics[width=0.4\textwidth]{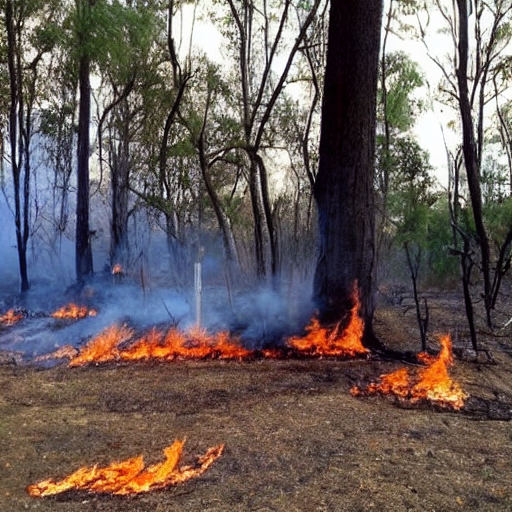}} \\ \hline

\textbf{GPT Evaluation Scores:} \\ 
\begin{tabular}{|p{4.3cm}|p{4.3cm}|p{4.3cm}|} 
\hline
\textbf{Criterion} & \textbf{Image 1 Score} & \textbf{Image 2 Score} \\ \hline
General Preference & 7 & 8 \\ \hline
Visual Appeal & 7 & 8 \\ \hline
Prompt Alignment & 6 & 8 \\ \hline
\end{tabular} \\ \hline

\textbf{GPT Justification:} The second image has a more visually appealing and convincing depiction of fire with clear, vibrant flames that have a more realistic quality. Compared to the first image, the fire in the second image matches the prompt more accurately, showing trees burning next to brush, whereas the first image emphasizes the brush more and the fire appears less integrated with trees. Overall, the artistic quality and prompt alignment are stronger in the second image. \\ \hline

\end{tabular}

\vspace{10pt}
\vskip -0.1in

\end{table}

\newpage
\begin{table}[h!]
\caption{Comparison of Stable Diffusion 1.5 fine-tuned using SFT and RLHF. Ratings and justification are based on GPT-4o evaluation.}
\vskip 0.15in
\centering
\renewcommand{\arraystretch}{1.5} 
\setlength{\tabcolsep}{10pt} 

\begin{tabular}{|p{15cm}|}
\hline
\multicolumn{1}{|c|}{\textbf{Text Prompt:} Candles on a windowsill.} \\ \hline

\textbf{Image 1: Stable Diffusion 1.5 (SFT) } \\ 
\makebox[\textwidth]{\includegraphics[width=0.4\textwidth]{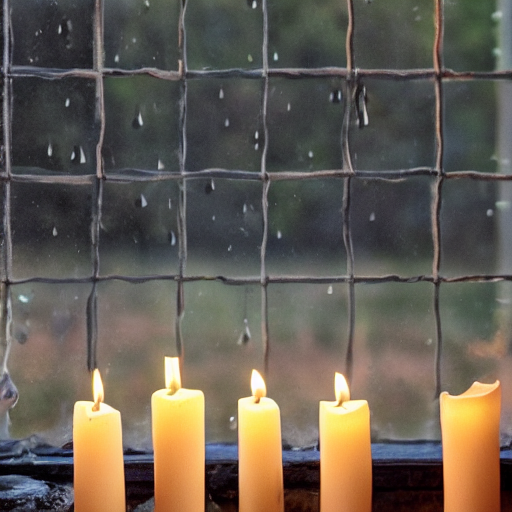}} \\ \hline

\textbf{Image 2: Stable Diffusion 1.5 + RLHF} \\
\makebox[\textwidth]{\includegraphics[width=0.4\textwidth]{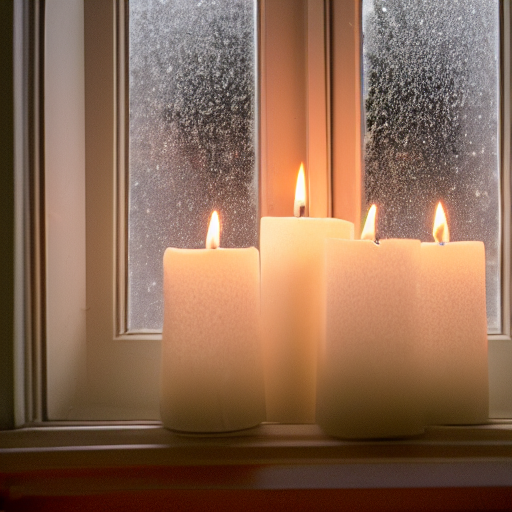}} \\ \hline

\textbf{GPT Evaluation Scores:} \\ 
\begin{tabular}{|p{4.3cm}|p{4.3cm}|p{4.3cm}|} 
\hline
\textbf{Criterion} & \textbf{Image 1 Score} & \textbf{Image 2 Score} \\ \hline
General Preference & 3 & 6 \\ \hline
Visual Appeal & 3 & 7 \\ \hline
Prompt Alignment & 4 & 7 \\ \hline
\end{tabular} \\ \hline

\textbf{GPT Justification:} The first image contains candles but lacks any visible flames, which are an essential part of fire generation. Consequently, it scores lower on all criteria. The second image features candles with visible flames, where the fire appears more realistic and aesthetically appealing. The flames in the second image are well-rendered and align better with the prompt, which specifically mentions candles on a windowsill. \\ \hline

\end{tabular}

\vspace{10pt}

\end{table}

\newpage
\begin{table}[h!]
\caption{Comparison of Stable Diffusion 1.5 fine-tuned using SFT and RLHF. Ratings and justification are based on GPT-4o evaluation.}
\vskip 0.15in
\centering
\renewcommand{\arraystretch}{1.5} 
\setlength{\tabcolsep}{10pt} 

\begin{tabular}{|p{15cm}|}
\hline
\multicolumn{1}{|c|}{\textbf{Text Prompt:} A large fire burning in the middle of a dry field.} \\ \hline

\textbf{Image 1: Stable Diffusion 1.5 (SFT) } \\ 
\makebox[\textwidth]{\includegraphics[width=0.4\textwidth]{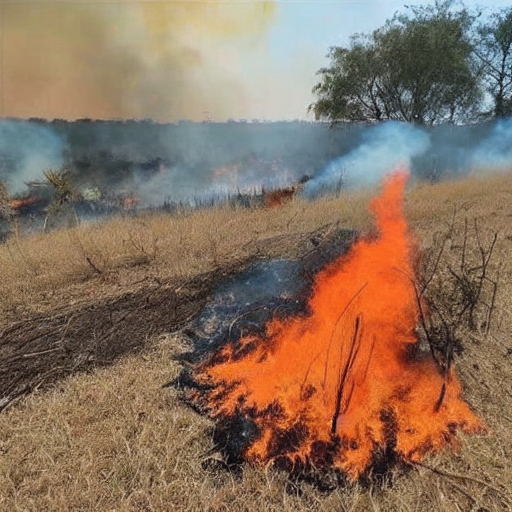}} \\ \hline

\textbf{Image 2: Stable Diffusion 1.5 + RLHF} \\
\makebox[\textwidth]{\includegraphics[width=0.4\textwidth]{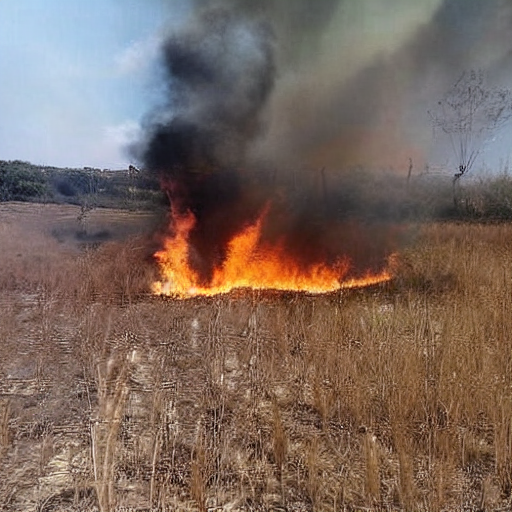}} \\ \hline

\textbf{GPT Evaluation Scores:} \\ 
\begin{tabular}{|p{4.3cm}|p{4.3cm}|p{4.3cm}|} 
\hline
\textbf{Criterion} & \textbf{Image 1 Score} & \textbf{Image 2 Score} \\ \hline
General Preference & 7 & 9 \\ \hline
Visual Appeal & 7 & 8 \\ \hline
Prompt Alignment & 6 & 8 \\ \hline
\end{tabular} \\ \hline

\textbf{GPT Justification:} The first image features a moderate fire with clearly visible orange flames and some surrounding smoke. The flames appear fairly realistic, but the intensity and overall impact are somewhat subdued.The second image shows a much more intense fire engulfing a tree, with vibrant orange flames and more dramatic smoke effects. This image is more visually striking and aligns better with the prompt of a large fire burning in a dry field.While both images depict fires in a dry field, the second one offers a more compelling and realistic representation of a large fire. \\ \hline

\end{tabular}

\vspace{10pt}

\end{table}

\newpage
\begin{table}[h!]
\caption{Comparison of Stable Diffusion 1.5 fine-tuned using SFT and RLHF. Ratings and justification are based on GPT-4o evaluation.}
\vskip 0.15in
\centering
\renewcommand{\arraystretch}{1.5} 
\setlength{\tabcolsep}{10pt} 

\begin{tabular}{|p{15cm}|}
\hline
\multicolumn{1}{|c|}{\textbf{Text Prompt:} The fire in the forest.} \\ \hline

\textbf{Image 1: Stable Diffusion 1.5 (SFT) } \\ 
\makebox[\textwidth]{\includegraphics[width=0.4\textwidth]{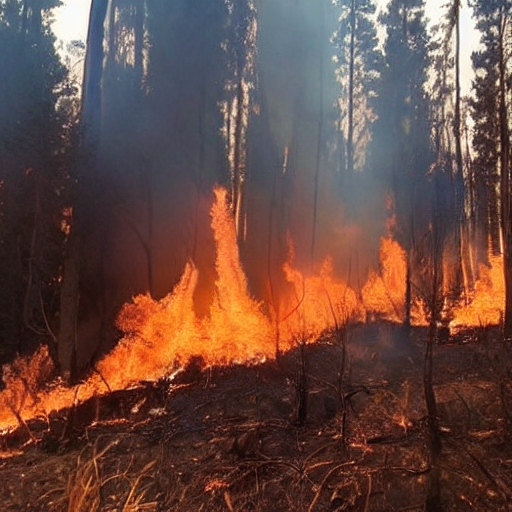}} \\ \hline

\textbf{Image 2: Stable Diffusion 1.5 + RLHF} \\
\makebox[\textwidth]{\includegraphics[width=0.4\textwidth]{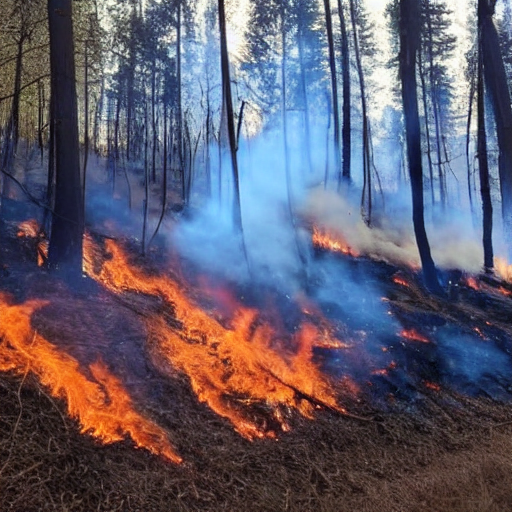}} \\ \hline

\textbf{GPT Evaluation Scores:} \\ 
\begin{tabular}{|p{4.3cm}|p{4.3cm}|p{4.3cm}|} 
\hline
\textbf{Criterion} & \textbf{Image 1 Score} & \textbf{Image 2 Score} \\ \hline
General Preference & 7 & 8 \\ \hline
Visual Appeal & 7 & 8 \\ \hline
Prompt Alignment & 8 & 9 \\ \hline
\end{tabular} \\ \hline

\textbf{GPT Justification:} The fire in the second image appears more vibrant and realistic compared to the first one. The flames in the second image have a stronger visual appeal with well-defined shapes and colors that enhance the artistic quality. Additionally, the fire in the second image aligns more accurately with the prompt, depicting a forest fire with a realistic and convincing appearance. The first image, while still good, has a slightly less detailed and less vibrant depiction of fire. \\ \hline

\end{tabular}

\vspace{10pt}

\end{table}

\newpage

\subsection{Qualitative Analysis in VLM Reasoning}
\label{sec:vlm_reasoning}
In this section, we present additional qualitative analyses that provide an in-depth, per-sample breakdown of correctness across the key components of fire reasoning.

First, we analyzed object match and environment match by comparing the outputs of the original model and the SFT fine-tuned model.

\begin{itemize}
    \item \textbf{Object Match:} 52.63\% of samples were correctly answered by both models, while 34.75\% were correctly answered only by the SFT model, and 10.10\% only by the original model. Only 2.53\% were incorrect in both cases.
    \item \textbf{Environment Match:} 62.73\% of cases were correct for both models, with 26.67\% correct only under SFT and 9.09\% only under the original model. Only 1.52\% of examples were incorrect in both.
\end{itemize}

These results demonstrate that the SFT model not only improves aggregate accuracy but also resolves additional cases that the original model fails to handle, supporting our claim that fine-tuning with DetectiumFire yields tangible gains in structured fire-related reasoning.

\begin{table}[h]
  \centering
  \caption{Normalized confusion matrices (\%) for four severity prediction.
  Rows are ground truth and columns are predictions.}
  \label{tab:confusion_fire}
  \vspace{0.5em}
  \resizebox{\textwidth}{!}{%
  \begin{tabular}{
      l
      S S S S
      @{\hspace{2em}}
      l
      S S S S
  }
    \toprule
    \multicolumn{5}{c}{\textbf{Original Model}} & \multicolumn{5}{c}{\textbf{SFT Model}} \\
    \cmidrule(lr){1-5} \cmidrule(lr){6-10}
    \multicolumn{1}{c}{\textbf{GT} $\downarrow$ / \textbf{Pred} $\rightarrow$} &
    \multicolumn{1}{c}{No Risk} &
    \multicolumn{1}{c}{Low} &
    \multicolumn{1}{c}{Moderate} &
    \multicolumn{1}{c}{High} &
    \multicolumn{1}{c}{\textbf{GT} $\downarrow$ / \textbf{Pred} $\rightarrow$} &
    \multicolumn{1}{c}{No Risk} &
    \multicolumn{1}{c}{Low} &
    \multicolumn{1}{c}{Moderate} &
    \multicolumn{1}{c}{High} \\
    \midrule
    No Risk   & 31.6 & 6.1 & 32.7 & 29.6 &
    No Risk   & 85.1 & 8.5 & 6.4  & 4.3 \\
    Low Risk  & 7.0  & 47.4 & 15.8 & 15.8 &
    Low Risk  & 1.8  & 66.7 & 10.5 & 7.0 \\
    Moderate  & 2.0  & 10.2 & 61.2 & 28.6 &
    Moderate  & 8.2  & 10.2 & 73.5 & 10.2 \\
    High Risk & 6.3  & 8.3  & 8.3  & 81.3 &
    High Risk & 2.1  & 2.1  & 12.5 & 87.5 \\
    \bottomrule
  \end{tabular}
  }
\end{table}

We further evaluated the models using a 4-class confusion matrix on fire severity prediction. Table~\ref{tab:confusion_fire} shows the normalized confusion matrices for both models. We observe that:

\begin{enumerate}
    \item All severity levels show clear improvements in diagonal accuracy (i.e., correct predictions) with the SFT model.
    \item False alarms are dramatically reduced. For instance, the original model misclassified "No Risk" cases as "High Risk" 30.6\% of the time, an unacceptable outcome in safety-critical applications. This is reduced to just 4.3\% in the SFT model.
    \item Improved severity alignment illustrates that the SFT model better understands fire context and threat levels.
\end{enumerate}

In addition, we provide detailed examples where the SFT model correctly identifies components of the fire scene that the original model fails to recognize.

\begin{table}[ht]

  \centering
  \caption{Example where the SFT model correctly identifies components of the fire scene that the original model fails to recognize.}
  \label{tab:qual_ex_1}
  \setlength{\tabcolsep}{6pt}\renewcommand{\arraystretch}{1.15}
  \begin{tabular}{@{}p{0.22\linewidth} p{0.75\linewidth}@{}}
    \toprule
    \multicolumn{2}{@{}c@{}}{%
      \includegraphics[width=\linewidth,height=0.25\textheight,keepaspectratio]{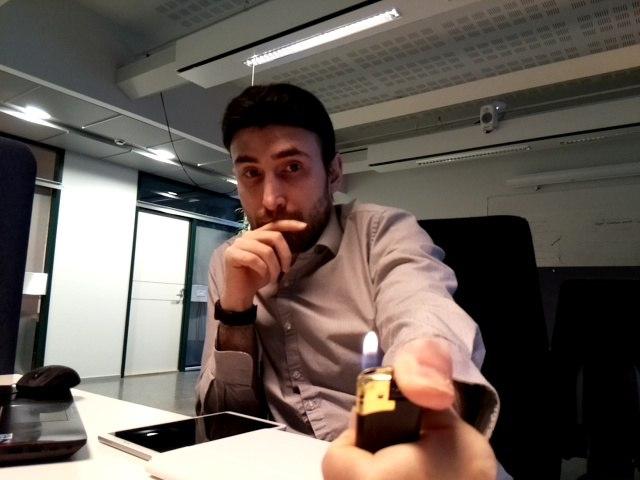}
    }\\
    \midrule
    \textbf{Original Output} &
    A man's shirt is burning, and the fire is spreading quickly, posing a moderate to severe risk to his safety and well-being. \\
    \textbf{SFT Output} &
    A cigarette lighter is being ignited in an indoor setting, with no immediate fire risk detected. \\
    \textbf{Justification} &
    The SFT model is correct; it recognizes a simple lighter ignition, whereas the original model incorrectly describes it as a severe fire incident. \\
    \bottomrule
  \end{tabular}

\end{table}

\begin{table}[ht]
  \centering
  \caption{Example where the SFT model correctly identifies components of the fire scene that the original model fails to recognize.}
  \label{tab:qual_ex_2}
  \setlength{\tabcolsep}{6pt}\renewcommand{\arraystretch}{1.15}
  \begin{tabular}{@{}p{0.22\linewidth} p{0.75\linewidth}@{}}
    \toprule
    \multicolumn{2}{@{}c@{}}{%
      \includegraphics[width=\linewidth,height=0.25\textheight,keepaspectratio]{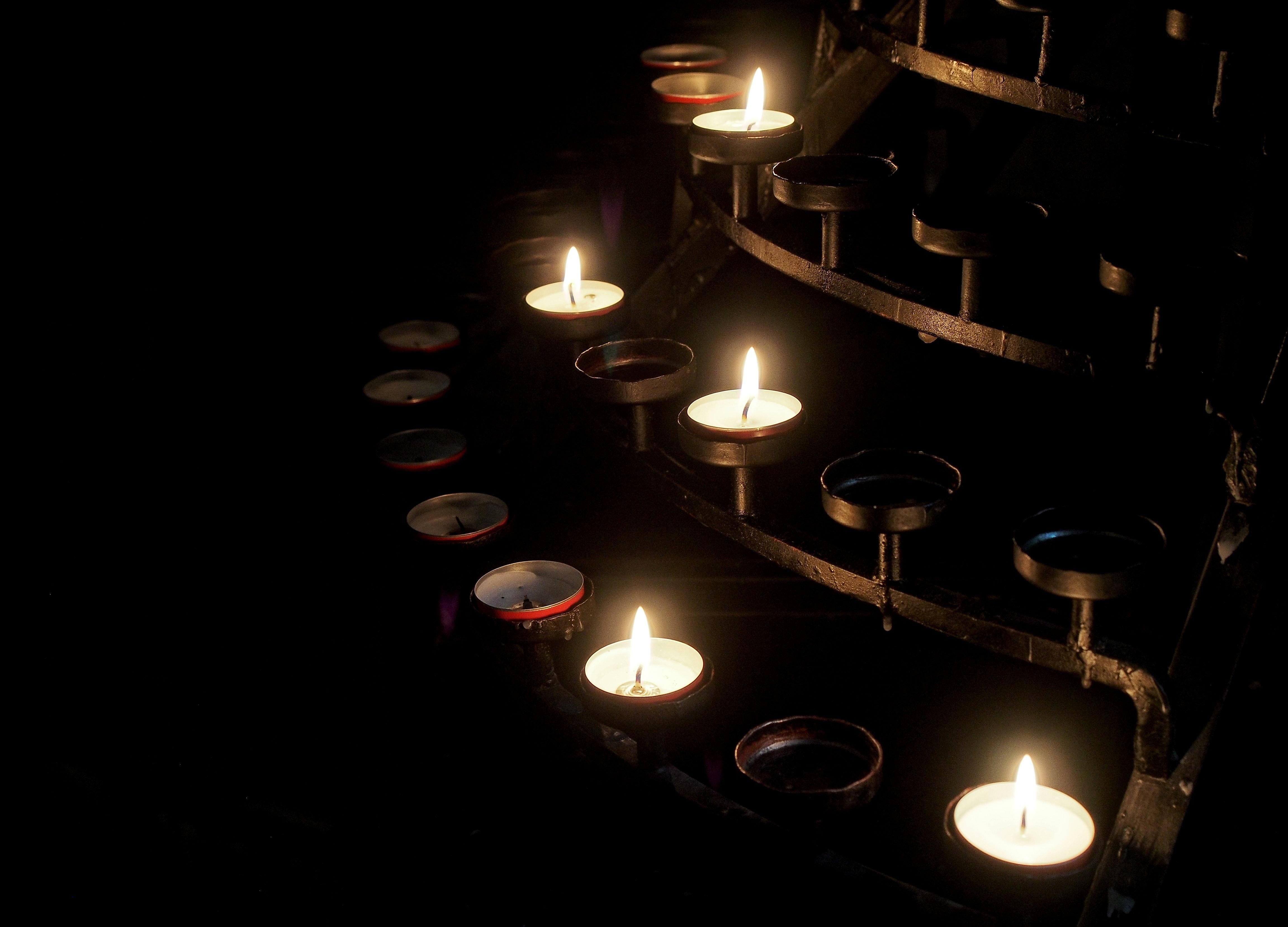}
    }\\
    \midrule
    \textbf{Original Output} &
    A moderate fire is burning candles in a dark room, with flames licking at the edges of the wax. \\
    \textbf{SFT Output} &
    Candles are burning on a metal stand indoors, with a minor severity. \\
    \textbf{Justification} &
    The SFT model accurately interprets the scene as a minor risk, whereas the original model exaggerates it as a moderate fire requiring attention.\\
    \bottomrule
  \end{tabular}
\end{table}

\begin{table}[ht]
  \centering
  \caption{Example where the SFT model correctly identifies components of the fire scene that the original model fails to recognize.}
  \label{tab:qual_ex_3}
  \setlength{\tabcolsep}{6pt}\renewcommand{\arraystretch}{1.15}
  \begin{tabular}{@{}p{0.22\linewidth} p{0.75\linewidth}@{}}
    \toprule
    \multicolumn{2}{@{}c@{}}{%
      \includegraphics[width=\linewidth,height=0.35\textheight,keepaspectratio, angle=90]{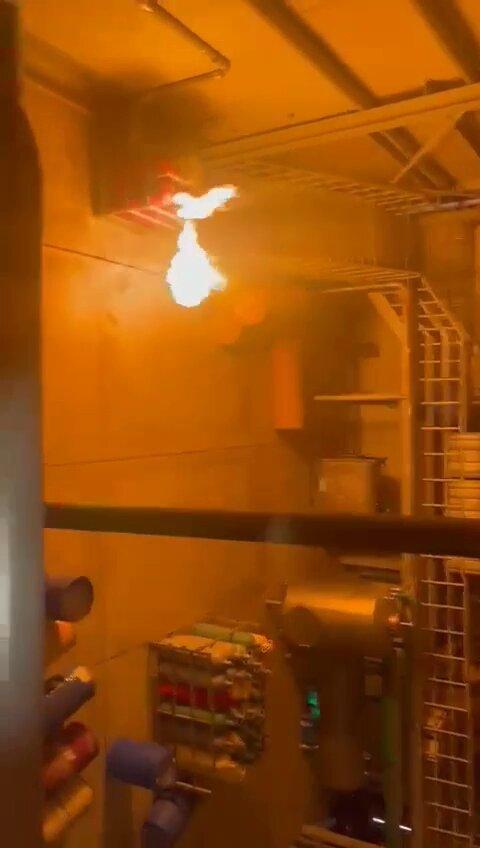}
    }\\
    \midrule
    \textbf{Original Output} &
    A moderate-sized fire is burning on the ceiling of a ship's deck, with a significant portion of the wooden planks and electrical wiring above the control room ablaze, posing a severe threat to the vessel's structural integrity and potentially causing a catastrophic explosion. \\
    \textbf{SFT Output} &
    A fire is burning on a pipe or ductwork in an industrial setting, with moderate severity. \\
    \textbf{Justification} &
    The SFT model correctly identifies the object and environment; the original model hallucinates nonexistent structure and exaggerates risk. \\
    \bottomrule
  \end{tabular}
\end{table}

\begin{table}[ht]
  \centering
  \caption{Example where the SFT model correctly identifies components of the fire scene that the original model fails to recognize.}
  \label{tab:qual_ex_4}
  \setlength{\tabcolsep}{6pt}\renewcommand{\arraystretch}{1.15}
  \begin{tabular}{@{}p{0.22\linewidth} p{0.75\linewidth}@{}}
    \toprule
    \multicolumn{2}{@{}c@{}}{%
      \includegraphics[width=\linewidth,height=0.25\textheight,keepaspectratio]{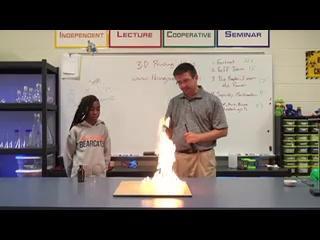}
    }\\
    \midrule
    \textbf{Original Output} &
    A moderate-sized fire is burning on a metal surface in a classroom, fueled by a chemical reaction between a metal and an acid, posing a significant risk to the students and teacher present in the room. \\
    \textbf{SFT Output} &
    A moderate fire is burning on a metal plate in a laboratory setting, with no immediate severity concerns. \\
    \textbf{Justification} &
    The SFT model correctly identifies the scene as a controlled experiment, while the original model misinterprets the setting and overstates the threat. \\
    \bottomrule
  \end{tabular}
\end{table}

We also include examples where the SFT model fails, highlighting the limitations of the current system and dataset. In summary, these results and examples illustrate that the SFT model fine-tuned on DetectiumFire not only improves numerical accuracy but also captures fire-specific scene understanding better than the original baseline. The reduced false alarms and more accurate scene interpretations highlight the utility of our dataset for real-world fire reasoning and support future work on tasks such as fire alarm triggering, threat assessment, and scene reporting. We also acknowledge areas for improvement and will address such cases in the next version of the dataset and models.

\begin{table}[h]
  \centering

  \caption{Example where the SFT model fails.}
  \label{tab:qual_ex_1_fail}
  \setlength{\tabcolsep}{6pt}\renewcommand{\arraystretch}{1.15}
  \begin{tabular}{@{}p{0.22\linewidth} p{0.75\linewidth}@{}}
    \toprule
    \multicolumn{2}{@{}c@{}}{%
      \includegraphics[width=\linewidth,height=0.25\textheight,keepaspectratio]{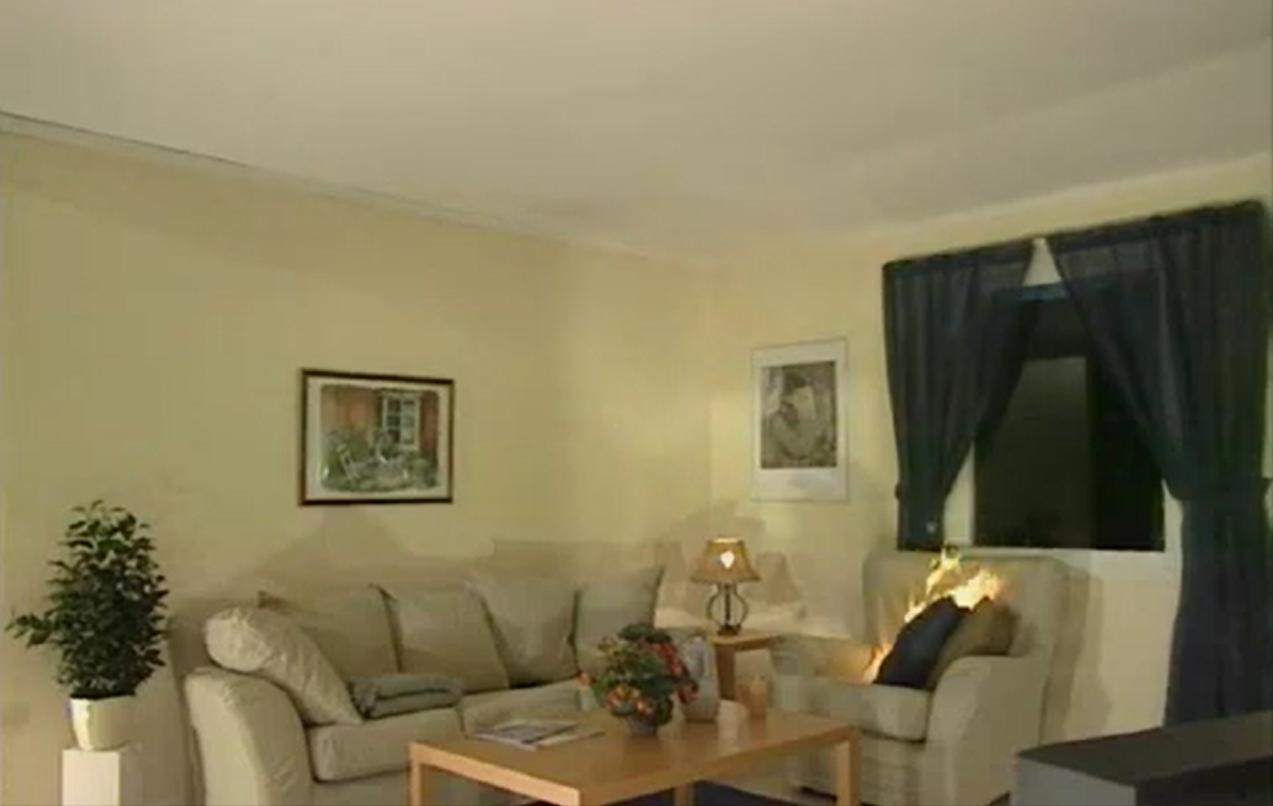}
    }\\
    \midrule
    \textbf{Original Output} &
   A moderate fire is burning the armchair and a corner of the wall in the living room. \\
    \textbf{SFT Output} &
    A moderate-sized fire is burning on a television set inside a room, with curtains partially covering the window. \\
    \textbf{Justification} &
    The SFT model fails to identify the correct object (armchair) and incorrectly predicts a TV fire. \\
    \bottomrule
  \end{tabular}

\end{table}

\begin{table}[h]
  \centering
  \caption{Example where the SFT model fails.}
  \label{tab:qual_ex_2_fail}
  \setlength{\tabcolsep}{6pt}\renewcommand{\arraystretch}{1.15}
  \begin{tabular}{@{}p{0.22\linewidth} p{0.75\linewidth}@{}}
    \toprule
    \multicolumn{2}{@{}c@{}}{%
      \includegraphics[width=\linewidth,height=0.25\textheight,keepaspectratio]{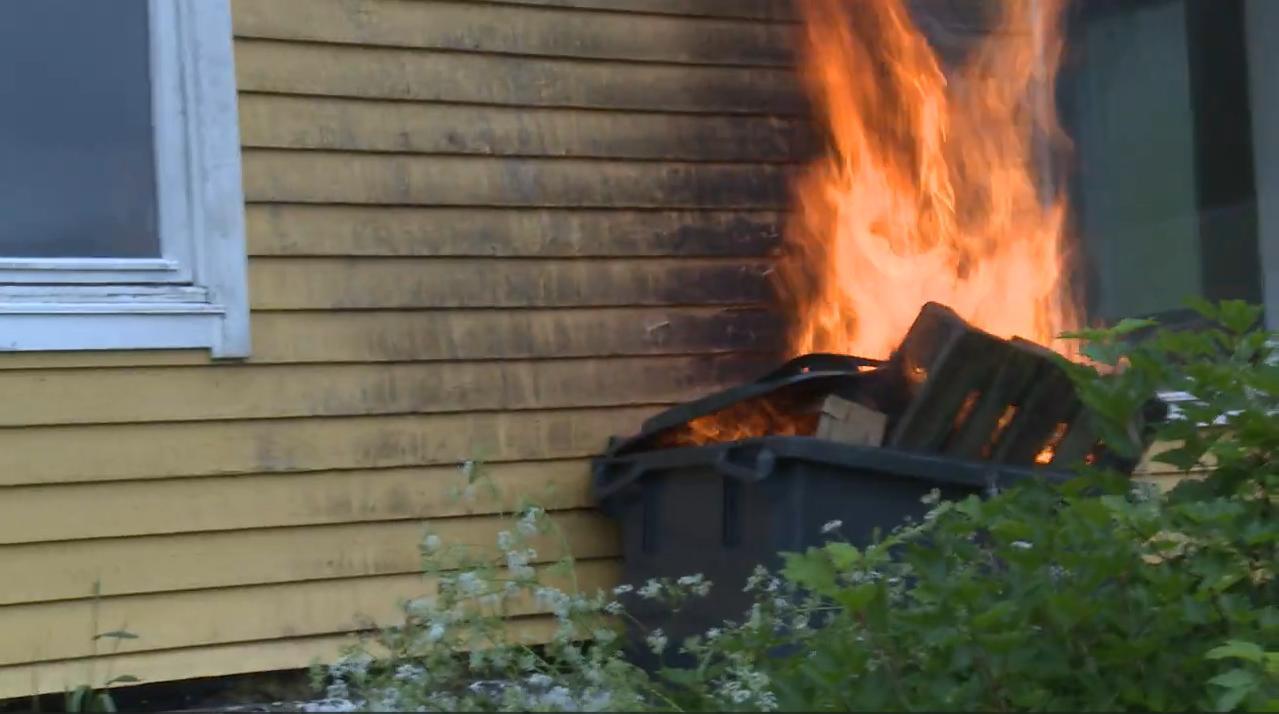}
    }\\
    \midrule
    \textbf{Original Output} &
    A moderate-sized fire is burning a pile of wooden pallets in a black plastic trash can, which is positioned against the side of a yellow house with horizontal wooden siding, likely causing significant damage to the structure. \\
    \textbf{SFT Output} &
    A moderate fire is burning on the exterior of a house, consuming wooden debris in a nearby container.\\
    \textbf{Justification} &
    The SFT model fails to detect the trash can as the burning object, providing a more generic description. \\
    \bottomrule
  \end{tabular}
\end{table}

\end{document}